\documentclass{article}

 \usepackage[main, final]{neurips_2026}
\usepackage[utf8]{inputenc} % allow utf-8 input
\usepackage[T1]{fontenc}    % use 8-bit T1 fonts
\usepackage{hyperref}       % hyperlinks
\usepackage{url}            % simple URL typesetting
\usepackage{booktabs}       % professional-quality tables
\usepackage{amsfonts}       % blackboard math symbols
\usepackage{nicefrac}       % compact symbols for 1/2, etc.
\usepackage{microtype}      % microtypography
\usepackage{xcolor}         % colors
\usepackage{graphicx}
\usepackage{amsmath}

\title{Flow Matching for Count Data}

\author{%
  Ganchao Wei\thanks{Corresponding author. Alternative email: \texttt{gwei@flatironinstitute.org}} \\
  Department of Neurobiology\\
  Department of Statistical Science\\
  Duke University\\
  Durham, NC, USA\\
  \texttt{weiganchao@gmail.com}
  \And
  John Pearson \\
  Department of Neurobiology\\
  Department of Electrical and Computer Engineering\\
  Duke University\\
  Durham, NC, USA\\
  \texttt{john.pearson@nyu.edu}
}

\begin{document}

\maketitle

\begin{abstract}
    High-dimensional count data arise in applications such as single-cell RNA sequencing and neural spike trains, where mappings between distributions across successive batches or time points form critical components of data analysis. The recent success of diffusion- and flow-based deep generative models for images, video, and text motivates extending these ideas to count-valued settings, but many existing methods either treat each count as a categorical state or transform counts into a continuous space, neither of which is natural or efficient when the count range is large. We propose count-FM, a flow-matching framework for count data based on a continuous-time birth-death process with local unit jumps. Count-FM learns marginal transitions efficiently in count space through simulation-free training of conditional transition rates, allowing transport between arbitrary count-distributed source and target populations. In simulation, count-FM variants achieve strong sample quality while using substantially fewer parameters. We further apply count-FM to scRNA-seq and neural spike-train data for unconditional generation, transport, and conditional generation. Across these tasks, count-FM yields improved sample quality, greater modeling efficiency, and interpretable transport paths.
\end{abstract}

\section{Introduction}
\label{intro}

High-dimensional count data arise in many scientific applications, including key biological data types such as single-cell RNA sequencing and neural spike trains. However, their discreteness, sparsity, and complex correlation structure make flexible joint modeling difficult, especially in high dimensions. Deep generative models provide a powerful framework for addressing this problem, with major developments including variational autoencoders \citep{kingma2014autoencoding}, generative adversarial networks \citep{goodfellow2014generative}, normalizing flows \citep{kingma2018glow}, diffusion models \citep{ho2020denoising,song2021scorebased}, and flow matching \citep{lipman2023flow}. In particular, diffusion- and flow-based methods have recently attracted substantial attention because of their strong empirical performance, flexible path-based formulations, and efficient training procedures \citep{ho2020denoising,song2021scorebased,lipman2023flow}. This progress has motivated growing interest in extending these generative frameworks from continuous domains to discrete data.

For modeling discrete data, recent approaches include categorical diffusion models such as D3PM \citep{austin2021structured}, tauLDR \citep{campbell2022a}, SEDD \citep{lou2024discrete}, and masked diffusion models \citep{sahoo2024simple,shi2024simplified,chao2025partial}, as well as discrete flow methods such as Dirichlet-FM \citep{stark2024dirichlet}, discrete-FM \citep{gat2024discrete}, and CTMC-based discrete flows \citep{campbell2024}. When adapted to count data, categorical-state methods typically require a predefined count range, and their categorical representation does not itself encode the ordering of counts. Another approach maps counts into a continuous space through dequantization or continuous representations \citep{ho2019flowpp,hoogeboom2020dequantization}, but this can blur the underlying discrete structure \citep{hoogeboom2019integer,luo2021graphdf}. Count-specific alternatives include Poisson-JUMP \citep{chen2023learning}, Count Bridges \citep{fishman2026countbridges}, and ItDPDM \citep{bhattacharya2025itdpdm}. Poisson-JUMP and ItDPDM are not designed for transport between arbitrary count distributions, while Count Bridges learns an endpoint posterior and exact finite-time bridges. 
% In contrast, count-FM directly learns local birth and death rates from closed-form conditional targets using a signed binomial bridge.

To address these challenges while retaining the efficient training and generation of flow matching, we propose count-FM, a flow-matching framework for count data based on a continuous-time birth-death process with local unit jumps. We model transport directly in count space through coordinate-wise local births and deaths, so that transitions respect the geometry of counts and remain count-valued along the path. The conditional binomial bridge yields an efficient training objective and enables tractable training of the marginal transition rates.
The model is also parameter-efficient, since it predicts only local birth and death rates rather than a full categorical distribution over all count levels, without requiring a predefined maximum count. This makes count-FM well suited to high-dimensional count data with large count ranges.

We first validate count-FM on a two-dimensional simulation before applying it to single-cell RNA-seq for unconditional generation and transport, followed by conditional generation of brain data, multiregion hippocampal and entorhinal spike trains. Our main contributions are as follows:
\begin{itemize}
    \item We propose count-FM, a flow-matching framework for count data based on a continuous-time birth-death process with local unit jumps, so that transport is modeled directly in count space and intermediate states remain count-valued along the path.
    \item We develop a tractable training scheme for marginal transition rates through a conditional binomial bridge while retaining the efficient training and generation of flow matching.
    \item We obtain a parameter-efficient formulation that predicts only local birth and death rates rather than a full categorical distribution over all count levels, making it well suited to high-dimensional count data with large count ranges.
\end{itemize}
The Python implementation of count-FM is available at \url{https://github.com/weigcdsb/count-flow-matching}.

\section{Method}
\label{method}

In this section, we first formulate count-FM and derive its training objective, then describe sampling, endpoint coupling, and conditional generation. Appendix~\ref{app:error_analysis} analyzes error from endpoint truncation, rate estimation, and sampling discretization.

Here, we consider a pair of counting distributions on $d$ variables, $\mathbb{N}_0^d = \{0,1,2,\ldots\}^d$, where $x_0 \sim p_0$ is a source sample and $x_1 \sim p_1$ is a target sample. We model the transition between these two distributions as a continuous-time Markov jump process (CTMC) \citep{campbell2022a, campbell2024} with local unit births and deaths. Let $X_t \in \mathbb{N}_0^d$ denote the count vector at time $t$, and let $e_i \in \mathbb{R}^d$ denote the $i$th standard basis vector, that is, the vector whose $i$th entry is $1$ and all other entries are $0$. Over a small time interval $h>0$, the process evolves according to
\[
\mathbb{P}(X_{t+h}=x+e_i \mid X_t=x)=h\,\lambda_{t,i}(x)+o(h),
\]
\[
\mathbb{P}(X_{t+h}=x-e_i \mid X_t=x)=h\,\mu_{t,i}(x)+o(h),
\]
and the probability of staying at $x$ is
\[
\mathbb{P}(X_{t+h}=x \mid X_t=x)
=
1-h\sum_{i=1}^d\bigl(\lambda_{t,i}(x)+\mu_{t,i}(x)\bigr)+o(h).
\]
Death is disallowed when $x^{(i)}=0$. The goal is to learn the time-dependent birth rates $\lambda_t(x)$ and death rates $\mu_t(x)$.

One benefit of this local birth-death parameterization is parameter efficiency for count-valued data.
Many categorical-state discrete diffusion and flow models for count data represent each coordinate over all count levels up to a maximum count, so the total output dimension is $\sum_{i=1}^d (C_i+1)$, where $C_i$ denotes the maximum count for coordinate $i$.
By contrast, count-FM parameterizes each coordinate using only birth and death rates for local $\pm1$ jumps, so its rate output dimension is always $2d$, regardless of maximum counts. This is more important in high-count and high-dimensional settings, where categorical-state parameterizations grow increasingly expensive.
Empirically, in both the simulation and scRNA experiments, the count-FM variants use fewer trainable parameters than all baseline models with available parameter counts (Tables~\ref{tab:toy_model_comp} and~\ref{tab:model_comp}).

\subsection{Conditional count bridge and training objective}
\label{sec:method_training}

Let $\pi$ be a coupling of $p_0$ and $p_1$. Given an endpoint pair $(x_0,x_1)\sim\pi$, we define the conditional bridge coordinatewise. For each coordinate $i$,
\begin{equation}
X_t^{(i)}
=
x_0^{(i)} + \operatorname{sgn}\!\bigl(x_1^{(i)}-x_0^{(i)}\bigr)\,B_t^{(i)},
\qquad
B_t^{(i)} \sim \mathrm{Binomial}\!\left(\left|x_1^{(i)}-x_0^{(i)}\right|,\, t\right),
\label{eq:conditional_binomial_bridge}
\end{equation}
% independently across coordinates given $(x_0,x_1)$, where $\operatorname{sgn}$ is the sign function. This bridge defines a conditional probability path $p_t(x\mid x_0, x_1)$
% whose coordinate-wise mean moves linearly from $x_0$ to $x_1$, with local count changes.
independently across coordinates given $(x_0,x_1)$, where $\operatorname{sgn}$ is the sign function. This bridge defines a conditional probability path $p_t(x\mid x_0, x_1)$ whose coordinate-wise mean moves linearly from $x_0$ to $x_1$, with local count changes. Conditional on $(x_0,x_1)$, each coordinate remains between its endpoints and evolves monotonically in expectation. After marginalization, however, the learned rates depend on the full state $x$, so generated trajectories need not be monotone and cross-coordinate dependence can be captured.

For a fixed coordinate \(i\), let \(x=X_t^{(i)}\) and write \(p_t^{(i)}(x\mid x_0,x_1)\) for the corresponding one-dimensional bridge marginal. Then mass preservation (equivalently, the one-dimensional Kolmogorov forward equation for a local CTMC \citep{holderrieth2025generator}) gives
\begin{align}
\partial_t p_t^{(i)}(x\mid x_0,x_1)
&=
\underbrace{
\lambda_{t,i}(x-1\mid x_0,x_1)\,
p_t^{(i)}(x-1\mid x_0,x_1)
}_{\text{birth inflow}}
\notag\\
&\quad+
\underbrace{
\mu_{t,i}(x+1\mid x_0,x_1)\,
p_t^{(i)}(x+1\mid x_0,x_1)
}_{\text{death inflow}}
\notag\\
&\quad-
\underbrace{
\bigl(
\lambda_{t,i}(x\mid x_0,x_1)
+
\mu_{t,i}(x\mid x_0,x_1)
\bigr)
p_t^{(i)}(x\mid x_0,x_1)
}_{\text{outflow from }x}.
\label{eq:mp_1d}
\end{align}
% \begin{align}
% \partial_t p_t^{(i)}(x\mid x_0,x_1)
% &=
% \lambda_{t,i}(x-1\mid x_0,x_1)\,p_t^{(i)}(x-1\mid x_0,x_1) \notag\\
% &+
% \mu_{t,i}(x+1\mid x_0,x_1)\,p_t^{(i)}(x+1\mid x_0,x_1) \notag\\
% &-
% \bigl(\lambda_{t,i}(x\mid x_0,x_1)+\mu_{t,i}(x\mid x_0,x_1)\bigr)p_t^{(i)}(x\mid x_0,x_1).
% \label{eq:mp_1d}
% \end{align}
This equation expresses local mass balance, where the change in mass equals influx minus outflux. Requiring the conditional rates to satisfy it ensures that the CTMC generates the conditional path $p_t(\cdot\mid x_0,x_1)$. Substituting the conditional binomial bridge \eqref{eq:conditional_binomial_bridge} into \eqref{eq:mp_1d} yields
$$
\lambda_{t,i}(x \mid x_0,x_1)
=
\frac{\bigl(x_1^{(i)}-x\bigr)_+}{1-t},
\qquad
\mu_{t,i}(x \mid x_0,x_1)
=
\frac{\bigl(x-x_1^{(i)}\bigr)_+}{1-t},
$$
where $(\cdot)_+$ indicates positive rectification. To avoid the singular endpoint at $t=1$, we truncate the path at $\tau:=1-\varepsilon_t$ and use the exact conditional rates above for $t\in[0,\tau]$.
The error analysis in Appendix~\ref{app:error_analysis} shows that this truncation contributes an $O(\varepsilon_t)$ error in total variation and an $O(\sqrt{\varepsilon_t})$ error in $W_2$, with the bounds depending on the source--target displacement induced by endpoint coupling $\pi$.
See Appendix~\ref{app:bridge_rates} for the detailed rate derivation.

The associated marginal transition rates $(\lambda_t,\mu_t)$ are obtained by averaging the conditional rates over endpoint pairs given the current state. By Proposition~1 of \citet{holderrieth2025generator}, they generate the marginal probability path
$
p_t(x)=\mathbb{E}_{(x_0,x_1)\sim\pi}\!\left[p_t(x\mid x_0,x_1)\right].
$
By Proposition~2 of \citet{holderrieth2025generator}, training with the conditional rates gives the same population gradients as training with the marginal rates. Thus, count-FM uses the same conditional-to-marginal matching principle as continuous flow matching, with transition rates replacing velocities. 
The CTMC formulation follows prior work \citep{campbell2022a,campbell2024}, while the signed binomial bridge and its closed-form rates are specific to count-FM.

To ensure zero death rate in any coordinate whose count is zero, we parameterize the model by nonnegative birth rates $\lambda_{\theta}(x,t)$ and death rates $\mu_{\theta}(x,t)=x \odot \beta_{\theta}(x,t)$,
where the death coefficients $\beta_{\theta}(x,t)$ are nonnegative. In the following experiments, birth and death rates are modeled using a single neural network with separate outputs.
We derive the training objective from the path-space Kullback--Leibler (KL) divergence induced by the conditional count bridge, following the standard path-space relative-entropy formulation for CTMCs \citep{arampatzis2015pathwise}. Let
$X_{(0,\tau]}:=\{X_s:0<s\le\tau\}$ and define
\begin{equation}
\mathcal{L}_{\mathrm{path}}(\theta)
=
\mathbb{E}_{(x_0,x_1)\sim\pi}
\left[
\mathrm{KL}\!\left(
p(X_{(0,\tau]}\mid X_0=x_0,X_1=x_1)
\;\middle\|\;
p_\theta(X_{(0,\tau]}\mid X_0=x_0)
\right)
\right].
\label{eq:path_kl_objective}
\end{equation}
For the local birth-death process, the path-space KL decomposes into local one-step KL terms. Taking the infinitesimal limit and dropping terms independent of $\theta$ gives the practical training objective
\begin{equation}
\mathcal{L}_{\mathrm{train}}(\theta)
=
\mathbb{E}_{\substack{(x_0,x_1)\sim\pi,\\
t\sim\mathrm{Unif}[0,\tau],\\
x\sim p_t(\cdot\mid x_0,x_1)}}
\left[
\sum_{i=1}^d
\ell\!\left(
\lambda_{t,i}(x\mid x_0,x_1),
\lambda_{\theta,i}(x,t)
\right)
+
\sum_{i=1}^d
\ell\!\left(
\mu_{t,i}(x\mid x_0,x_1),
\mu_{\theta,i}(x,t)
\right)
\right],
\label{eq:rate_matching_objective}
\end{equation}
where $\ell(u,v)=v-u\log v$. We use $\log(v+\varepsilon_{\ell})$ for numerical stability. Appendices~\ref{app:poisson_loss} and~\ref{app:path_kl_interp} show that
$
\mathcal{L}_{\mathrm{path}}(\theta)
=
\tau\mathcal{L}_{\mathrm{train}}(\theta)+c,
$
where $c$ does not depend on $\theta$. Thus the path-space and practical objectives are equivalent for optimization. This is analogous to the data-augmentation view of diffusion objectives in \citet{kingma2023understanding}. Appendix~\ref{app:error_analysis} further relates the rate-matching error to the final distribution error.

\subsection{Sample generation}
\label{sec:method_sampling}

After training, we generate samples by simulating the learned birth-death process forward from $t=0$ to $t=\tau=1-\varepsilon_t$. The initial state is sampled from the source distribution $p_0$, which can be either a simple count-valued distribution (for unconditional generation) or the observed source population.
Furthermore, we use a first-order local-jump discretization. Like tau-leaping methods for Markov jump processes \citep{gillespie2001approximate}, the sampler freezes the transition rates within each time step, while restricting each coordinate to at most one jump.
With $K$ uniform steps and step size $\Delta=\tau/K=(1-\varepsilon_t)/K$, define the total jump rate for coordinate $i$ at state $x$ and time $t$ as
\[
r_i(x,t)=\lambda_{\theta,i}(x,t)+\mu_{\theta,i}(x,t).
\]
For each coordinate $i$ with $r_i(x,t)>0$, we sample one of three outcomes,
\[
p_i^{\mathrm{stay}}=\exp(-r_i\Delta),\qquad
p_i^{\mathrm{birth}}=\bigl(1-\exp(-r_i\Delta)\bigr)\frac{\lambda_{\theta,i}}{r_i},\qquad
p_i^{\mathrm{death}}=\bigl(1-\exp(-r_i\Delta)\bigr)\frac{\mu_{\theta,i}}{r_i}.
\]
When $r_i(x,t)=0$, we set $p_i^{\mathrm{stay}}=1$ and $p_i^{\mathrm{birth}}=p_i^{\mathrm{death}}=0$.
The next state is obtained by applying the sampled unit update in each coordinate. Repeating this update over the $K$ time steps from $t=0$ to $t=\tau$ yields an approximate sample trajectory. 

% \textcolor{red}{
Appendix~\ref{app:error_analysis} decomposes the terminal distribution error into 1) endpoint truncation, 2) rate estimation, and 3) sampling discretization. The truncation bound favors a smaller endpoint gap (smaller $\varepsilon_t$) and couplings with shorter source--target displacements. The squared total variation error due to rate estimation is at most half the gap between the optimal and training losses. The discretization bound further differentiates between errors due to multiple jumps within each step and from rate freezing, favoring lower jump intensities ($\sum_i^dr_i(x,t)$) and rates that vary slowly in time and state. Under bounded and Lipschitz rates, the local discretization error is $O(\Delta^2)$, and the global error is $O(K^{-1})$. Our implementation, the signed binomial bridge, partly follows this guidance. Its monotone conditional trajectories minimize the number of unit jumps between fixed endpoints, with constant expected jump intensity equal to the source-target $L_1$ displacement. The OT coupling in Section~\ref{sec:method_ot} can also lower the expected jump intensity if it reduces the mean source-target $L_1$ displacement relative to independent coupling.
% }

% Appendix~\ref{app:error_analysis} separates the discretization error from the at-most-one-jump approximation and rate freezing. Under bounded and Lipschitz learned rates, the local error is $O(\Delta^2)$ and the final discretization error is $O(K^{-1})$. Therefore, to keep the discretization error small, smaller time steps should be used when jump rates are large or vary rapidly.

\subsection{Endpoint coupling}
\label{sec:method_ot}

The endpoint coupling $\pi(x_0,x_1)$ determines how source and target samples are paired during training and thus influences the learned marginal transition path. Here, we either consider independent coupling, $\pi_{\mathrm{ind}}(x_0,x_1)=p_0(x_0)p_1(x_1)$, (\textbf{count-FM}) or a minibatch optimal-transport (OT) coupling (\textbf{count-FM-OT}), similar to OT-CFM \citep{tong2024improving}.
For the latter, we draw source and target minibatches $\{x_{0,b}\}_{b=1}^B$ and $\{x_{1,b}\}_{b=1}^B$. These define empirical measures $\hat p_0^B=\frac{1}{B}\sum_{b=1}^B\delta_{x_{0,b}}$ and $\hat p_1^B=\frac{1}{B}\sum_{b=1}^B\delta_{x_{1,b}}$, both with uniform weights $u_B=\frac{1}{B}\mathbf{1}_B$. We then compute the empirical OT coupling
\[
\Gamma^\star
=
\arg\min_{\Gamma\in\Pi(u_B,u_B)}
\sum_{b=1}^B\sum_{b'=1}^B \Gamma_{bb'}\,c(x_{0,b},x_{1,b'}),
\]
where $\Pi(u_B,u_B)$ denotes the set of joint distributions on the two minibatches with marginals $u_B$. We sample endpoint pairs $(x_0,x_1)$ from $\Gamma^\star$ and then use the same conditional bridge construction and training objective as in Section~\ref{sec:method_training}. This minibatch OT step can be viewed as an empirical approximation to a population coupling between $p_0$ and $p_1$. We use the symmetric Poisson cost
\[
c(x,y)
=
\sum_{i=1}^d
\left[
x_i\log\frac{x_i+\varepsilon_c}{y_i+\varepsilon_c}
+
y_i\log\frac{y_i+\varepsilon_c}{x_i+\varepsilon_c}
\right],
\]
where $\varepsilon_c>0$ is a small constant for numerical stability. This is a numerically stabilized symmetrized divergence between count vectors.

In the experiments, we report both count-FM and count-FM-OT for unconditional generation. 
% For the conditional generation task, we use the independently coupled version only.
The OT coupling changes the endpoint coupling, and hence modifies the geometry of the learned marginal transport path. In practice, OT coupling tends to induce transitions with lower curvature, which is beneficial for interpretation. It can also improve sampling efficiency, in the sense that under the same discretization scheme, similar sample quality may be achieved with fewer function evaluations (NFEs) or less wall-clock time. 
We verify this empirically in Appendix Figure~\ref{fig:ot_efficiency}, where the OT-coupled version generally reaches a given quality level more quickly than the independently coupled version.

\subsection{Conditional generation}
\label{sec:method_conditional}

For conditional generation, we augment the model with covariates $y$ and use conditional rates $\lambda_{\theta}(x,t,y)$ and $\mu_{\theta}(x,t,y)$.
We adapt classifier-free guidance (CFG) \citep{ho2021classifierfree} to the transition rates. During training, conditioning variables are randomly dropped, allowing a single network to learn both conditional and unconditional rates..
This avoids training a separate auxiliary classifier and provides a simple way to control the strength of conditioning at sampling time through the guidance scale.
At sampling time, we combine the conditional and unconditional rates as
\[
\lambda_{\theta}^{\mathrm{cfg}}
=
\Bigl[
\lambda_{\theta}^{\mathrm{uncond}}
+
w\bigl(\lambda_{\theta}^{\mathrm{cond}}-\lambda_{\theta}^{\mathrm{uncond}}\bigr)
\Bigr]_+,
\qquad
\mu_{\theta}^{\mathrm{cfg}}
=
\Bigl[
\mu_{\theta}^{\mathrm{uncond}}
+
w\bigl(\mu_{\theta}^{\mathrm{cond}}-\mu_{\theta}^{\mathrm{uncond}}\bigr)
\Bigr]_+,
% \lambda_{\theta}^{\mathrm{cfg}}
% =
% \lambda_{\theta}^{\mathrm{uncond}}
% +
% w\Bigl(\lambda_{\theta}^{\mathrm{cond}}-\lambda_{\theta}^{\mathrm{uncond}}\Bigr),
% \qquad
% \mu_{\theta}^{\mathrm{cfg}}
% =
% \mu_{\theta}^{\mathrm{uncond}}
% +
% w\Bigl(\mu_{\theta}^{\mathrm{cond}}-\mu_{\theta}^{\mathrm{uncond}}\Bigr),
\]
where $w\ge0$ is the guidance scale and $[\cdot]_+$ ensures nonnegative rates. The operator $[\cdot]_+$ has no effect for $0\le w\le1$, but may alter the extrapolated rates when $w>1$.
% where $w\ge 0$ is the guidance scale. 
Here, $w=0$ recovers unconditional generation, $0<w<1$ interpolates between unconditional and conditional generation, and $w=1$ gives the standard conditional model. 
Values $w>1$ heuristically extrapolate beyond the standard conditional model, which can improve condition alignment and sharpen condition-specific structure.
% Values $w>1$ strengthen the influence of the conditioning variables beyond the standard conditional model, which can improve condition alignment and sharpen condition-specific structure. 
At the same time, overly large $w$ may distort the marginal distribution and reduce diversity.

\section{Simulation}
\label{sim}

We begin with a two-dimensional example simulation designed to evaluate both sample quality and intermediate transport behavior.
The target distribution is an equal-weight mixture of two Gamma-Poisson components with modes near $(60,5)$ and $(60,40)$. For a fair comparison, all models except Poisson-JUMP \citep{chen2023learning} use the same discrete-uniform source distribution on a square count grid covering the displayed data range, while Poisson-JUMP uses its own Poisson-based source construction. 
% We compare both count-FM and count-FM-OT against several representative discrete generative baselines: Dirichlet-FM (cite) and discrete-FM (cite) from discrete flow matching, D3PM (cite), tauLDR (cite), and SEDD (cite) from discrete diffusion-style modeling, and Poisson-JUMP as a count-specific jump-process baseline. Most of these baselines are based on categorical-state parameterizations, so their intermediate transitions are not naturally aligned with the quantitative geometry of count data and can therefore be less smooth or interpretable in count space. 
We compare both count-FM and count-FM-OT against several representative discrete generative baselines: Dirichlet-FM \citep{stark2024dirichlet} and discrete-FM \citep{gat2024discrete} from discrete flow matching, D3PM \citep{austin2021structured}, tauLDR \citep{campbell2022a}, and SEDD \citep{lou2024discrete} from discrete diffusion-style modeling, and Poisson-JUMP, a count-specific jump model. Most of these baselines are based on categorical-state parameterizations. Poisson-JUMP instead works directly with counts, but unlike count-FM it does not learn an explicit conditional bridge with local birth and death rates.
For count-FM, we use a small time-varying MLP backbone with hidden width 32, with shared hidden layers and separate output channels for the birth and death rates. For the competing methods, we use standard or matched-capacity architectures. 

We evaluate sample quality using the 1- and 2-Wasserstein distances, $W_1$ and $W_2$, and $\mathrm{MMD}^2_{\mathrm{RBF}}$, the squared maximum mean discrepancy with a Gaussian RBF kernel \citep{gretton12a}. For each model, we use 2{,}000 generated samples under a common evaluation seed and report the mean and standard deviation across target resamples in Table~\ref{tab:toy_model_comp}. count-FM-OT ranks first under both $W_1$ and $W_2$. SEDD obtains the lowest $\mathrm{MMD}^2_{\mathrm{RBF}}$. Thus, the main conclusion is robust to the choice of Wasserstein order, while both count-FM variants use substantially fewer trainable parameters than the competing models.

Figure~\ref{fig:sim_snapshot} in Appendix~\ref{sim_appendix} shows representative intermediate samples along each model's native sampling trajectory. Figure~\ref{fig:sim_model_bridge} further compares the marginal bridges under a common progress variable, with Poisson-JUMP excluded because no analogous bridge is available. Under this normalization, both count-FM variants evolve smoothly in count space, while count-FM-OT follows a visibly straighter transition path. In contrast, the categorical-state diffusion- and flow-based baselines move most mass toward the target early along the common progress scale, indicating a more abrupt transition in count space. This geometric advantage is also reflected in sampling efficiency. Appendix Figure~\ref{fig:ot_efficiency} shows that count-FM-OT generally reaches a given $W_2$ or $\mathrm{MMD}^2_{\mathrm{RBF}}$ level with fewer NFEs or less runtime than the independently coupled version.
\begin{table}[h!]
  \caption{\textbf{Performance comparison on the two-dimensional simulation.}
  For each model, 2{,}000 samples are generated under a common evaluation seed, and results are reported as mean $\pm$ standard deviation across target resamples. All models except Poisson-JUMP use the same discrete-uniform source distribution. Model names include the number of trainable parameters in parentheses. Best and second-best results are bolded and underlined, respectively.}
  \label{tab:toy_model_comp}
  \centering
  \small
  \begin{tabular}{lccc}
    \toprule
    Model (params)
    & $W_1 \downarrow$
    & $W_2 \downarrow$
    & $\mathrm{MMD}^2_{\mathrm{RBF}} \downarrow$ \\
    \midrule
    count-FM-OT (2,372)
    & \textbf{1.685 $\pm$ 0.163}
    & \textbf{2.341 $\pm$ 0.330}
    & \underline{0.0005 $\pm$ 0.0004} \\

    count-FM (2,372)
    & 1.866 $\pm$ 0.219
    & \underline{2.583 $\pm$ 0.378}
    & 0.0009 $\pm$ 0.0004 \\

    discrete-FM (14,250)
    & 1.942 $\pm$ 0.205
    & 2.684 $\pm$ 0.363
    & 0.0007 $\pm$ 0.0005 \\

    D3PM (15,306)
    & 1.982 $\pm$ 0.190
    & 2.812 $\pm$ 0.446
    & 0.0010 $\pm$ 0.0006 \\

    SEDD (15,306)
    & \underline{1.747 $\pm$ 0.115}
    & 3.051 $\pm$ 0.165
    & \textbf{0.0002 $\pm$ 0.0003} \\

    Poisson-JUMP (67,330)
    & 2.974 $\pm$ 0.271
    & 3.794 $\pm$ 0.394
    & 0.0035 $\pm$ 0.0009 \\

    tauLDR (15,306)
    & 2.417 $\pm$ 0.292
    & 3.855 $\pm$ 0.586
    & 0.0024 $\pm$ 0.0012 \\

    Dirichlet-FM (15,306)
    & 3.569 $\pm$ 0.379
    & 6.101 $\pm$ 0.450
    & 0.0035 $\pm$ 0.0014 \\
    \bottomrule
  \end{tabular}
\end{table}

\section{Applications}
\label{app}

\subsection{Single-cell RNA-seq generation and transport}
\label{app:scRNA}

We study the Dentate Gyrus scRNA-seq dataset \citep{Hochgerner2018DentateGyrus}, which contains 2,930 cells and 13,913 genes, together with cell-type and developmental-age annotations spanning multiple lineages. The two developmental time points used in our transport experiment are postnatal day 12 (P12, 1,124 cells) and postnatal day 35 (P35, 1,806 cells). We consider two tasks: unconditional generation from the marginal distribution and transport from P12 to P35 along development.

\subsubsection{Unconditional generation}
\label{app:scRNA_uncond_gen}

For unconditional generation, each model is trained to generate full gene-expression count vectors from the marginal cell distribution. We use a random 80/20 train-test split over all 2,930 cells, giving 2,344 training cells and 586 held-out test cells. We compare count-FM and count-FM-OT with representative baselines for single-cell generation, including scVI \citep{Lopez2018DeepGenerative} as a latent-variable baseline, scDiffusion \citep{btae518} and scLDM \citep{palla2025a} as latent diffusion models, and DCM \citep{bhattacharya2026discrete} and CFGen \citep{palma2025multimodal} as additional recent generative baselines for scRNA count data. 
For both scRNA experiments, count-FM-(OT) uses a shared transformer backbone with hidden dimension 256, depth 8, and 8 attention heads, with separate output channels for the birth and death rates. The competing methods use their implementations with comparable size settings for a fair comparison.

We evaluate held-out sample quality using $W_2$ and $\mathrm{MMD}^2_{\mathrm{RBF}}$ computed in a PCA feature space built from the top 2{,}000 variable genes after normalization and log transformation, with the number of PCs chosen to explain 90\% of the variance. Here, $W_2$ is computed between Gaussian fits to the real and generated PCA features. The results are summarized in Table~\ref{tab:model_comp}. Count-FM achieves the best performance on both metrics, while count-FM-OT performs comparably. As in the simulation experiment (Section~\ref{sim}), OT endpoint coupling improves sampling efficiency at lower budgets, with count-FM-OT reaching comparable quality using fewer NFEs or less runtime, while the gap becomes smaller at larger budgets (Appendix Figure~\ref{fig:ot_efficiency}). 
Paired comparisons across the five matched seeds show that count-FM improves both metrics over all five competing methods after Holm correction \citep{holm1979simple}, while an additional marker-gene analysis shows that count-FM achieves the lowest marker-expression MAE, indicating the most accurate recovery of cell-type-relevant expression levels (Appendix Tables~\ref{tab:scrna_paired} and~\ref{tab:scrna_marker}).
Count-FM uses 45.70\% fewer trainable parameters than scVI and 5.46\% fewer than scLDM.
% Among the models with available parameter counts, count-FM and count-FM-OT achieve strong held-out sample quality with a substantially more efficient parameterization.
\begin{table}[h!]
  \caption{\textbf{Performance comparison across generative models on the Dentate Gyrus dataset for unconditional generation.} Results are reported as mean $\pm$ standard deviation over 5 repeated runs. Model names include the number of trainable parameters in parentheses when available. count-FM achieves the best performance on both $W_2$ and $\mathrm{MMD}^2_{\mathrm{RBF}}$, while count-FM-OT performs comparably.}
  \label{tab:model_comp}
  \centering
  \begin{tabular}{lcc}
    \toprule
    Model (params) & $W_2 \downarrow$ & $\mathrm{MMD}^2_{\mathrm{RBF}} \downarrow$ \\
    \midrule
    count-FM (9,857,792)        & \textbf{20.456 $\pm$ 0.055} & \textbf{0.0185 $\pm$ 0.0005} \\
    count-FM-OT (9,857,792)     & 20.553 $\pm$ 0.054 & 0.0195 $\pm$ 0.0007 \\
    scLDM (10,426,955)           & 20.739 $\pm$ 0.018 & 0.0202 $\pm$ 0.0003 \\
    scDiffusion (34,704,897)    & 20.545 $\pm$ 0.051 & 0.0225 $\pm$ 0.0004 \\
    scVI (18,155,426)            & 20.687 $\pm$ 0.042 & 0.0229 $\pm$ 0.0002 \\
    
    DCM (39,848,665)            & 23.566 $\pm$ 0.010 & 0.0296 $\pm$ 0.0000 \\
    CFGen (29,165,380)          & 22.226 $\pm$ 0.106 & 0.0361 $\pm$ 0.0008 \\
    \bottomrule
  \end{tabular}
\end{table}

\subsubsection{Transport from P12 to P35}
\label{app:scRNA_12_to_35}

We next study transport between postnatal days 12  (P12) and 35 (P35) in the Dentate Gyrus dataset. 
A key advantage of count-FM is that it defines transport directly in count space, with local $\pm1$ transitions that preserve the ordered count geometry and make intermediate changes gradual and interpretable.
% A key advantage of count-FM is that it defines transport directly in count space, so the intermediate states remain count-valued samples and are therefore interpretable as meaningful transitional distributions.
This is especially useful in developmental settings, where one would like to inspect not only the endpoints but also the full transition path. Related single-cell methods have likewise emphasized reconstructing developmental trajectories and fate structure, for example through optimal transport and fate-mapping frameworks \citep{Schiebinger2019, Lange2022}.

The Dentate Gyrus data have a dominant developmental trajectory along the granule-cell lineage, while several mature populations form comparatively stable side lineages. In particular, prior analyses \citep{Cui2024} of this dataset identify the main progression from neuroblast cells to granule immature cells and then to granule mature cells, whereas mature side populations evolve separately rather than along the main granule trajectory \citep{Hochgerner2018DentateGyrus}. 
To avoid source-target pairings driven only by geometric proximity, we transport P12 count profiles to P35 count profiles using lineage-restricted OT couplings. Specifically, OT matching is performed only within lineage-consistent source-target groups, so the learned transport respects known developmental structure and avoids biologically implausible transitions across incompatible cell states. This restriction encourages straighter, lower-curvature paths that are easier to interpret biologically. We use separate 80/20 train-test splits for each subset, stratified by cell cluster to preserve cluster composition.

\begin{figure}[h!]
  \centering
  \includegraphics[width=\linewidth]{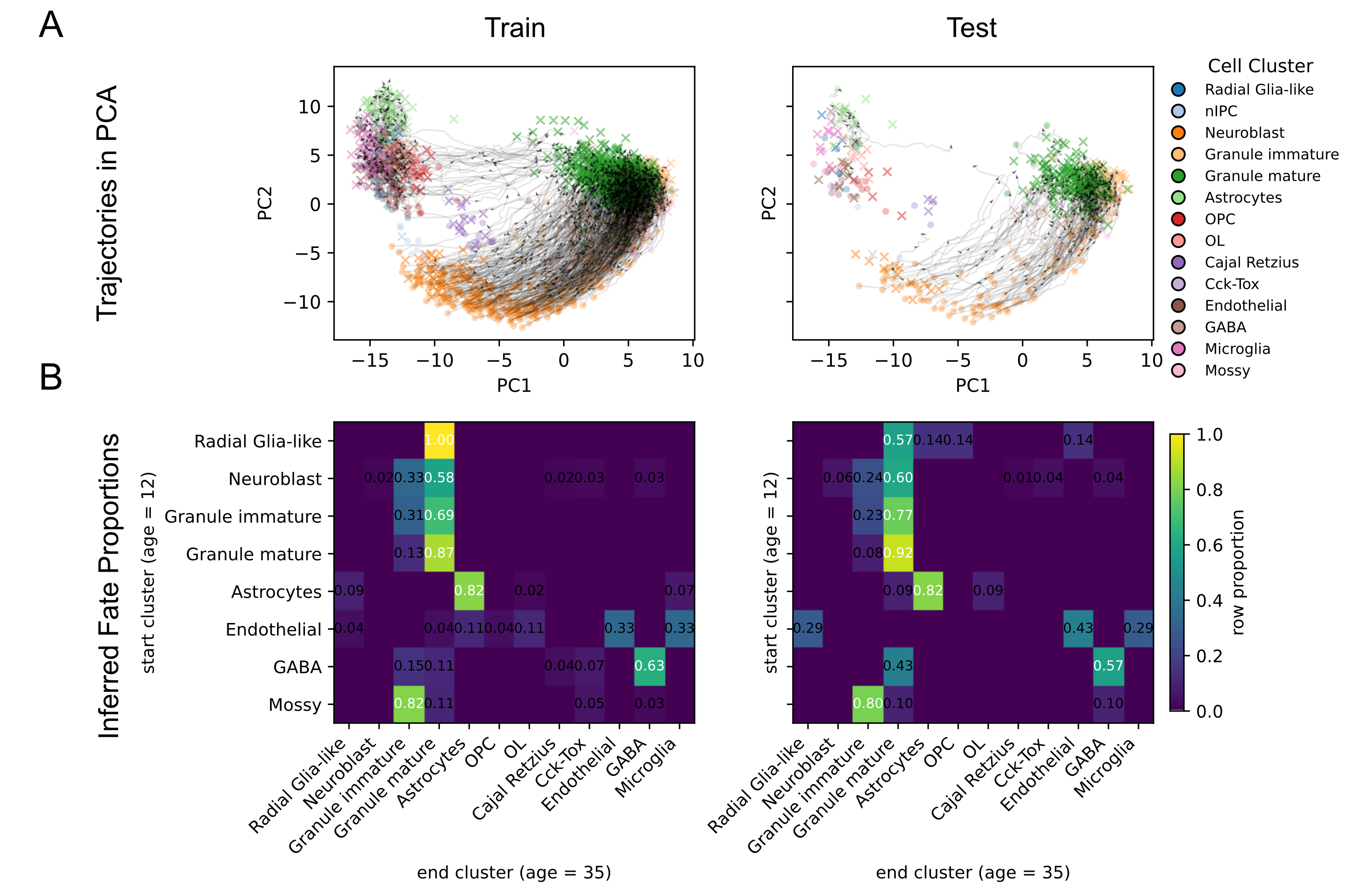}
  \caption{\textbf{Transport from P12 to P35.} \textbf{A.} Generated transport trajectories in PCA space for the training and test splits. Circles denote the P12 source cells and crosses denote the P35 target cells, with colors indicating cell cluster identity. Black curves show representative generated trajectories, which smoothly connect P12 to P35. \textbf{B.} Inferred fate proportions under P12-to-P35 transport. Rows correspond to source clusters at P12 and columns to inferred endpoint clusters at P35. Entries are row-normalized proportions for the training and test splits.}
  \label{fig:scRNA_12_35_transfer}
\end{figure}

Figure~\ref{fig:scRNA_12_35_transfer} summarizes the P12-to-P35 transport results. In panel A, generated trajectories move smoothly from the P12 cells toward the P35 cells, with colors indicating cell lineage. In panel B, the inferred terminal states are largely lineage-consistent. Neuroblast and granule immature cells predominantly map to granule immature or granule mature states, while mature side populations such as astrocytes, endothelial, and GABA cells mostly remain within their own lineages.
Because lineage-restricted OT incorporates known lineage information, this lineage consistency is partly induced by the coupling rather than discovered by the model.
Appendix Figures~\ref{fig:scRNA_snapshot_train} and~\ref{fig:scRNA_snapshot_test} further show that the transition unfolds progressively over time on both the training and test sets. 
Together, these results illustrate how count-FM can be used to model structured developmental transport directly in count space.

\subsection{Conditional generation on neural spike counts}
\label{sec:hc3_main}

\begin{figure}[h!]
  \centering
  \includegraphics[width=\linewidth]{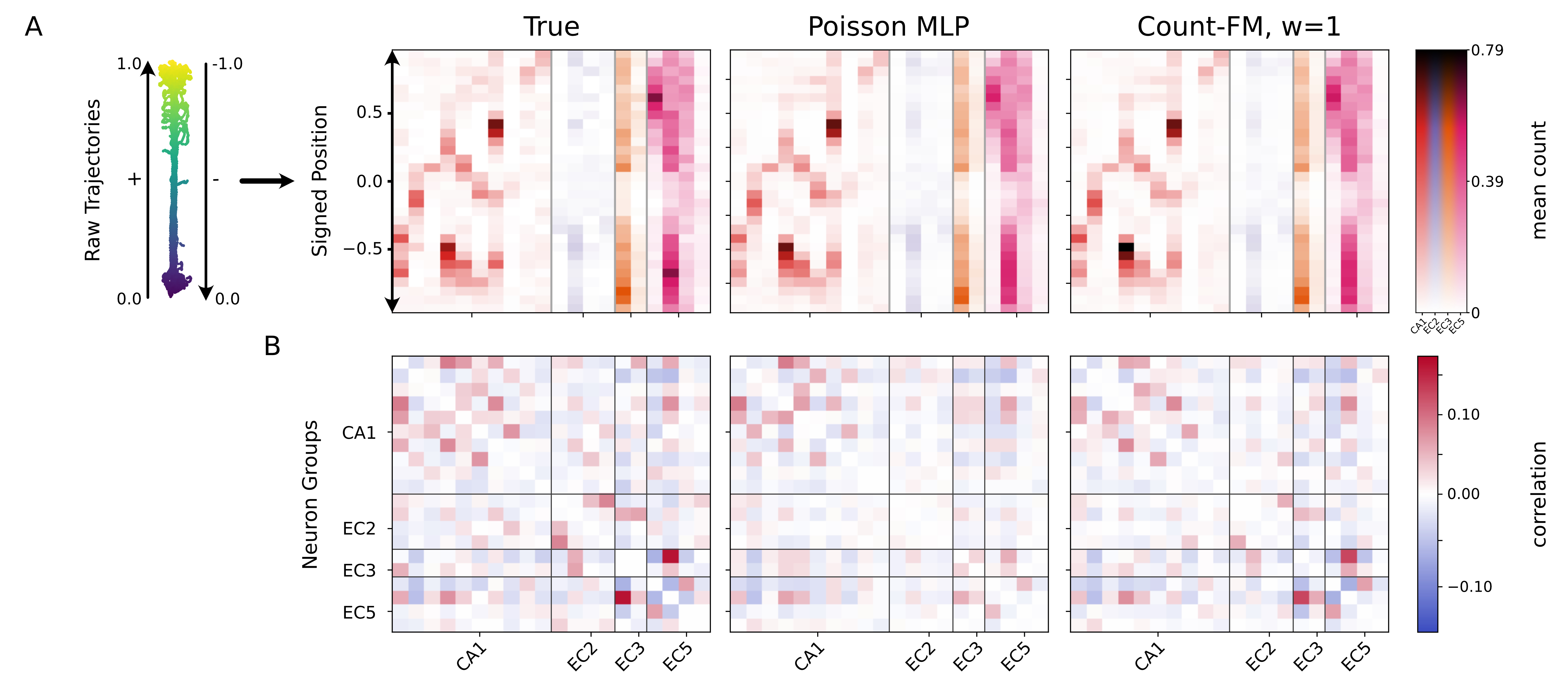}
  \caption{\textbf{Conditional generation on the hc-3 linear-track session.} Summaries are estimated from 100 generated samples per held-out covariate. \textbf{A.} Signed position is constructed from linearized position and running direction. Mean-response heatmaps show bin-wise mean counts for the true held-out data, Poisson MLP, and count-FM with $w=1$. \textbf{B.} Population correlation matrices for the same active-neuron set. Count-FM better preserves cross-neuron dependence, while Poisson MLP underestimates population correlations.}
  \label{fig:hc_3_main}
\end{figure}

We next study conditional generation of neural spike counts from a large collection of multi-region hippocampal and entorhinal cortex recordings in behaving Long-Evans rats. In our experiment, we focus on one linear-track session (\texttt{ec013.719}) from the CRCNS hc-3 dataset \citep{Mizuseki2013HC3,Mizuseki2014Neurosharing}, retaining 62 simultaneously recorded units from CA1, EC2, EC3, and EC5, yielding 29{,}239 observations after preprocessing. 
We use an 80/20 train-test split, giving 23{,}391 training observations and 5{,}848 held-out observations. The covariates are linearized position and running direction, combined into a single  signed position (Figure~\ref{fig:hc_3_main}A). 
Here, in contrast to generative benchmarking (simulation, Section~\ref{sim}) and unconditional generation (scRNA-seq, Section~\ref{app:scRNA_uncond_gen}), we focus on conditional count modeling versus standard regression baselines. Specifically, we compare count-FM with guidance scales $w\in\{0,1,2\}$ against two reference models, a deterministic MLP regressor, which predicts only conditional means, and a Poisson MLP baseline, which models count noise but has limited ability to capture complex correlation structure. Count-FM instead models a complicated joint conditional count distribution and supports classifier-free guidance (CFG), which controls the strength of conditional information without training an auxiliary classifier, as described in Section~\ref{sec:method_conditional}. For a fair comparison, all models use comparable width-128 three-layer MLP backbones with SELU activations. The MLP and Poisson MLP therefore serve as diagnostic baselines for conditional mean and covariance rather than a comprehensive generative benchmark.

Figure~\ref{fig:hc_3_main} shows representative results for location-modulated neurons from these regions. Many recorded neurons in hippocampal and entorhinal regions are silent or only weakly location-specific, including many interneuron-like units \citep{ThompsonBest1989,Epsztein2011,Hangya2010}, so we display a subset of neurons from each region to make the spatial response pattern visible. In the mean-response heatmaps (Figure~\ref{fig:hc_3_main}A and Appendix Figure~\ref{fig:hc3_mean_heatmap}), all models recover the main spatial response patterns. These response patterns are well known in classic findings that hippocampal neurons encode position and can be modulated by running direction \citep{McNaughton1983,Markus1995}. However, the models differ more clearly in their population dependence structure (Figure~\ref{fig:hc_3_main}B). Count-FM with $w=1$ better preserves the observed cross-neuron correlation pattern, whereas the Poisson MLP underestimates the dependence structure.

% We further examine the effect of guidance scale in Appendix~\ref{app:hc3_appendix}. Count-FM with $w=0$ has no conditional information and therefore only captures marginal firing-rate differences across neurons. Increasing the guidance scale to $w=2$ strengthens the location-specific response pattern and increases contrast, but the resulting mean response is less well calibrated to the true data. For the dependence structure (Appendix Figure~\ref{fig:hc3_dependence}), the MLP mean regressor is not included because it is not a generative count model. The unconditional count-FM model ($w=0$) still captures some marginal population dependence, whereas the Poisson MLP substantially underestimates the correlation structure. Stronger guidance again amplifies the dependence pattern (e.g., $w=2$), but sacrifices calibration to the data, consistent with the discussion in Section~\ref{sec:method_conditional}.
We further examine the effect of guidance scale in Appendix~\ref{app:hc3_appendix}. For mean responses (Appendix Figure~\ref{fig:hc3_mean_heatmap}), count-FM with $w=0$ lacks conditional information and only captures marginal firing-rate differences, while $w=2$ sharpens location-specific responses but reduces calibration to the true data. For population correlation structure (Appendix Figure~\ref{fig:hc3_dependence}), the MLP mean regressor is excluded because it is not a generative count model. Count-FM with $w=0$ still captures some marginal population dependence, Poisson MLP substantially underestimates correlations, and $w=2$ amplifies dependence at the cost of calibration, consistent with the discussion in Section~\ref{sec:method_conditional}.

To quantify these qualitative observations, we report several held-out metrics in Table~\ref{tab:hc3_conditional}. For a signed-position bin $b$ and neuron $j$, let $\mu_{bj}$ denote the held-out mean count and $\hat\mu_{bj}$ the model mean, estimated from 50 generated samples for generative models. We evaluate
$\mathrm{RMSE}_{\mu}
=
\sqrt{
\frac{\sum_b n_b \sum_j (\hat\mu_{bj}-\mu_{bj})^2}
{\sum_b n_b d}
}$
and analogously compute $\mathrm{RMSE}_{\mathrm{var}}$ and $\mathrm{RMSE}_{0}$ using the bin-wise variance and zero fraction. To evaluate correlation structure, we compute the off-diagonal covariance matrix $C_b$ of active neurons within each bin and calculate
$\mathrm{Cov}_{\mathrm{F}}
=
\frac{\sum_b n_b \lVert \hat C_b-C_b\rVert_F}{\sum_b n_b}$,
where $\hat C_b$ is the corresponding off-diagonal covariance matrix computed from generated samples in bin $b$. We also summarize place-field sharpness by the raw contrast
$\mathrm{contrast}_j
=
\frac{
\max_b \hat\mu_{bj}
-
\operatorname{median}_b(\hat\mu_{bj})
}
{\operatorname{mean}_b(\hat\mu_{bj})}$,
averaged over active neurons. Larger contrast indicates stronger peak-to-background tuning.

Table~\ref{tab:hc3_conditional} summarizes results over 5 replicated runs, each using a different random seed for train-test splitting and model retraining. Count-FM with $w=1$ gives the strongest overall generative performance, with the lowest RMSE$_\mu$, RMSE$_{\mathrm{var}}$, RMSE$_0$, and Cov$_{\mathrm{F}}$, indicating best fit to mean firing rate, sparsity, and correlation structure. The Poisson MLP baseline remains competitive for the mean response, but its variance, zero-fraction, and covariance errors are substantially larger. The unconditional model ($w=0$) has no spatial specificity. Increasing guidance to $w=2$ produces the sharpest place fields, but worsens most fidelity metrics, indicating over-sharpening. Overall, $w=1$ gives the best balance between spatial tuning and distributional calibration.

In a separate dataset, we also evaluate conditional generation on piriform cortex odor-response spike trains in Appendix~\ref{app:odor_spike}, where count-FM shows a similar guidance tradeoff. The mean MLP and Poisson MLP baselines recover only coarse response structure, with overly smooth or diffuse responses. In contrast, count-FM better captures odor- and respiration-dependent response patterns. Moderate guidance (e.g., $w=1$) gives better calibration, while stronger guidance (e.g., $w=2$) sharpens the pattern but can over-amplify spike counts.
\begin{table}[h!]
  \caption{\textbf{Conditional generation on the hc-3 linear-track session.} Results are mean $\pm$ standard deviation across 5 random seeds. count-FM with $w=1$ gives the best performance on mean firing rate (RMSE$_\mu$), sparsity (RMSE$_0$), and covariance structure (RMSE$_{\mathrm{var}}$ and Cov$_{\mathrm{F}}$), while $w=2$ gives the largest contrast but worsens calibration.}
  \label{tab:hc3_conditional}
  \centering
  \small
  \setlength{\tabcolsep}{4pt}
  \begin{tabular}{lccccc}
    \toprule
    Model & RMSE$_\mu \downarrow$ & RMSE$_{\mathrm{var}} \downarrow$ & RMSE$_0 \downarrow$ & Cov$_{\mathrm{F}} \downarrow$ & Contrast \\
    \midrule
    MLP mean        & 0.026 $\pm$ 0.002 & --                & --                & --                & 3.087 $\pm$ 0.034 \\
    Poisson MLP     & 0.027 $\pm$ 0.002 & 0.115 $\pm$ 0.003 & 0.027 $\pm$ 0.001 & 0.564 $\pm$ 0.012 & 3.134 $\pm$ 0.102 \\
    count-FM, $w=0$ & 0.070 $\pm$ 0.002 & 0.110 $\pm$ 0.005 & 0.043 $\pm$ 0.000 & 0.451 $\pm$ 0.021 & 0.109 $\pm$ 0.007 \\
    count-FM, $w=1$ & \textbf{0.026 $\pm$ 0.002} & \textbf{0.048 $\pm$ 0.005} & \textbf{0.016 $\pm$ 0.001} & \textbf{0.344 $\pm$ 0.017} & 3.209 $\pm$ 0.086 \\
    count-FM, $w=2$ & 0.061 $\pm$ 0.002 & 0.094 $\pm$ 0.007 & 0.036 $\pm$ 0.001 & 0.451 $\pm$ 0.020 & \textbf{4.521 $\pm$ 0.076} \\
    \bottomrule
  \end{tabular}
\end{table}

\section{Discussion}
\label{discussion}

In this work, we introduced count-FM, a flow-matching framework for count data based on local birth-death dynamics. By modeling transport through local births and deaths directly in count space, count-FM respects the underlying count geometry while avoiding the large categorical-state output parameterizations used by many existing discrete generative models. In simulation and in applications to scRNA-seq and neural spike trains, count-FM achieved high sample quality with fewer parameters, while its intermediate paths remain in count space and evolve through local changes, supporting their interpretation as source-to-target transformations.

Although count-FM performed well across our experiments, the model still has several limitations and natural directions for improvement. First, the conditional binomial bridge is deliberately simple, but it may not be stochastic enough to adequately explore plausible intermediate paths. This may increase extrapolation error and make the learned source-to-target transition overly rigid. When different futures are indistinguishable from the current state, marginal rate matching averages their conditional dynamics, and additional conditioning is needed to recover fate-specific transitions. A natural extension is to replace the binomial bridge with a more flexible stochastic bridge, for example a beta-binomial variant, or more generally a latent stochastic path construction in the spirit of stochastic interpolants \citep{albergo2023si,albergo2025stochastic} and GP-CFM \citep{wei2025gpcfm}.

Second, our current local-jump sampler is first-order and can introduce larger discretization error when jump rates are large or rapidly varying. Although model size is independent of the count range, large source-target displacements require more sampling steps. A source distribution closer to the target count scale or OT coupling can reduce typical displacements and therefore the required number of steps. The sampler itself could also be improved using nonuniform or adaptive time grids, uniformization-based CTMC simulation \citep{chen2025uniformization}, more accurate bridge-aware simulation schemes \citep{hobolth2009endpoint}, and inference-time correction methods \citep{skreta2025fkc}. Accelerated transition operators based on shortcut models \citep{frans2025shortcut} or inner-flow sampling \citep{holderrieth2026glass} may further reduce the required number of sampling steps, while flow-map learning \citep{boffi2025flowmap,potaptchik2026discreteflowmaps} can bypass stepwise simulation and remove this source of discretization error.

In summary, we introduced count-FM, a flow-matching framework for count data based on local birth-death dynamics. Across simulation and applications to biomedical data, including scRNA-seq and neural spike trains, count-FM achieved strong sample quality with fewer parameters than competing categorical-state approaches, while preserving count-valued intermediate paths. These results suggest that count-FM provides an effective framework for modeling high-dimensional count data by enabling generation and transport directly in the count space.

\begin{ack}
This work was supported by a grant from the National Institutes of Health (RF1DA056376) to JP through the BRAIN Initiative.
\end{ack}

% \section*{References}

\bibliographystyle{plainnat}
\bibliography{count_fm}

%%%%%%%%%%%%%%%%%%%%%%%%%%%%%%%%%%%%%%%%%%%%%%%%%%%%%%%%%%%%

\appendix

\section{Method details and derivations}
\label{app:method_derivations}

This appendix provides derivations and technical details for the construction in Section~\ref{method}. We first derive the conditional bridge rates, then derive the local rate-matching objective and its path-space KL interpretation. We finally analyze the endpoint truncation, rate-estimation, and sampling-discretization errors of count-FM.

\subsection{Derivation of conditional bridge rates}
\label{app:bridge_rates}

This subsection derives the conditional birth and death rates used in Section~\ref{sec:method_training}. We work in one dimension and suppress the coordinate index $i$. Let $x_0,x_1 \in \mathbb{N}_0$ and define the bridge
\[
X_t
=
x_0 + \operatorname{sgn}(x_1-x_0)\,B_t,
\qquad
B_t \sim \mathrm{Binomial}\!\left(|x_1-x_0|, t\right).
\]

For fixed endpoints $(x_0,x_1)$, let
\[
p_t(x\mid x_0,x_1)=\mathbb{P}(X_t=x\mid x_0,x_1)
\]
denote the one-dimensional conditional bridge pmf. Then mass preservation is
\begin{align}
\partial_t p_t(x\mid x_0,x_1)
&=
\lambda_t(x-1\mid x_0,x_1)\,p_t(x-1\mid x_0,x_1)
+
\mu_t(x+1\mid x_0,x_1)\,p_t(x+1\mid x_0,x_1) \notag\\
&\quad-
\bigl(\lambda_t(x\mid x_0,x_1)+\mu_t(x\mid x_0,x_1)\bigr)p_t(x\mid x_0,x_1).
\label{eq:mp_1d_app}
\end{align}
This is the one-dimensional special case of the Kolmogorov forward equation (KFE) for CTMCs \citep{campbell2022a}.

\paragraph{Case 1: $x_1 \ge x_0$.}
In this case, the bridge is increasing, with
\[
p_t(x\mid x_0,x_1)
=
\binom{x_1-x_0}{x-x_0}
t^{\,x-x_0}(1-t)^{\,x_1-x},
\qquad x \in \{x_0,\ldots,x_1\},
\]
and $\mu_t(x\mid x_0,x_1)=0$ for $0\le t\le 1$. Hence mass preservation reduces to
\begin{equation}
\partial_t p_t(x\mid x_0,x_1)
=
\lambda_t(x-1\mid x_0,x_1)\,p_t(x-1\mid x_0,x_1)
-
\lambda_t(x\mid x_0,x_1)\,p_t(x\mid x_0,x_1).
\label{eq:mp_pos}
\end{equation}
Differentiating the binomial pmf gives
\[
\partial_t p_t(x\mid x_0,x_1)
=
\left(
\frac{x-x_0}{t}
-
\frac{x_1-x}{1-t}
\right)p_t(x\mid x_0,x_1),
\]
and
\[
\frac{p_t(x-1\mid x_0,x_1)}{p_t(x\mid x_0,x_1)}
=
\frac{x-x_0}{x_1-x+1}\cdot\frac{1-t}{t}.
\]
Let
\[
\lambda_t(x\mid x_0,x_1)=\frac{x_1-x}{1-t},
\]
then
\[
\lambda_t(x-1\mid x_0,x_1)\,p_t(x-1\mid x_0,x_1)
=
\frac{x-x_0}{t}\,p_t(x\mid x_0,x_1),
\]
and hence \eqref{eq:mp_pos} holds. Therefore, for $x_1 \ge x_0$,
\[
\lambda_t(x\mid x_0,x_1)=\frac{x_1-x}{1-t},
\qquad
\mu_t(x\mid x_0,x_1)=0.
\]

\paragraph{Case 2: $x_1 < x_0$.}
In this case, the bridge is decreasing, with
\[
p_t(x\mid x_0,x_1)
=
\binom{x_0-x_1}{x_0-x}
t^{\,x_0-x}(1-t)^{\,x-x_1},
\qquad x \in \{x_1,\ldots,x_0\},
\]
and $\lambda_t(x\mid x_0,x_1)=0$ for $0\le t\le 1$. Mass preservation reduces to
\begin{equation}
\partial_t p_t(x\mid x_0,x_1)
=
\mu_t(x+1\mid x_0,x_1)\,p_t(x+1\mid x_0,x_1)
-
\mu_t(x\mid x_0,x_1)\,p_t(x\mid x_0,x_1).
\label{eq:mp_neg}
\end{equation}
Applying the same calculation as in Case 1 to the reduced mass-preservation equation \eqref{eq:mp_neg} gives
\[
\mu_t(x\mid x_0,x_1)=\frac{x-x_1}{1-t},
\qquad
\lambda_t(x\mid x_0,x_1)=0.
\]

Combining the two cases yields
\[
\lambda_t(x\mid x_0,x_1)=\frac{(x_1-x)_+}{1-t},
\qquad
\mu_t(x\mid x_0,x_1)=\frac{(x-x_1)_+}{1-t}.
\]
Applying this coordinatewise gives the formula in Section~\ref{sec:method_training}.

\subsection{Local KL derivation of the rate-matching training objective}
\label{app:poisson_loss}

This subsection derives the rate-matching training objective \eqref{eq:rate_matching_objective} in Section~\ref{sec:method_training} from the infinitesimal KL divergence between the target and model one-step transition probabilities.

Fix a bridge state $x$ at time $t$. For a small step size $h>0$, define the target local transition distribution $q_h(\cdot\mid x,x_0,x_1)$ by
\begin{align*}
q_h(x+e_i\mid x,x_0,x_1)
&=
h\,\lambda_{t,i}(x\mid x_0,x_1)+o(h), \notag\\
q_h(x-e_i\mid x,x_0,x_1)
&=
h\,\mu_{t,i}(x\mid x_0,x_1)+o(h),
\end{align*}
and
\[
q_h(x\mid x,x_0,x_1)
=
1-h\sum_{i=1}^d\bigl(\lambda_{t,i}(x\mid x_0,x_1)+\mu_{t,i}(x\mid x_0,x_1)\bigr)+o(h).
\]
The model for local transition probability $q_h^{\theta}(\cdot\mid x,t)$ is defined analogously by replacing $\lambda_{t,i}(x\mid x_0,x_1),\mu_{t,i}(x\mid x_0,x_1)$ with $\lambda_{\theta,i}(x,t),\mu_{\theta,i}(x,t)$.

Let
\[
a_t(x\mid x_0,x_1)=\sum_{i=1}^d\bigl(\lambda_{t,i}(x\mid x_0,x_1)+\mu_{t,i}(x\mid x_0,x_1)\bigr),
\qquad
a_t^{\theta}(x,t)=\sum_{i=1}^d\bigl(\lambda_{\theta,i}(x,t)+\mu_{\theta,i}(x,t)\bigr).
\]
We consider the local KL divergence
\[
\mathrm{KL}\!\left(
q_h(\cdot\mid x,x_0,x_1)
\;\middle\|\;
q_h^{\theta}(\cdot\mid x,t)
\right).
\]
% We consider the local KL divergence
% \[
% D_{\mathrm{KL}}\!\bigl(q_h(\cdot\mid x,x_0,x_1)\,\Vert\,q_h^{\theta}(\cdot\mid x,t)\bigr).
% \]

For the no-jump event,
\[
q_h(x\mid x,x_0,x_1)\log\frac{q_h(x\mid x,x_0,x_1)}{q_h^{\theta}(x\mid x,t)}
=
\bigl(1-h\,a_t(x\mid x_0,x_1)+o(h)\bigr)
\log\frac{1-h\,a_t(x\mid x_0,x_1)+o(h)}{1-h\,a_t^{\theta}(x,t)+o(h)}.
\]
As $h\to 0$, using $\log(1+u)=u+o(u)$ as $u\to 0$, and noting that $q_h(x\mid x,x_0,x_1)=1+O(h)$,
\[
q_h(x\mid x,x_0,x_1)\log\frac{q_h(x\mid x,x_0,x_1)}{q_h^{\theta}(x\mid x,t)}
=
h\bigl(a_t^{\theta}(x,t)-a_t(x\mid x_0,x_1)\bigr)+o(h).
\]

For the jump events,
\[
q_h(x+e_i\mid x,x_0,x_1)\log\frac{q_h(x+e_i\mid x,x_0,x_1)}{q_h^{\theta}(x+e_i\mid x,t)}
=
h\,\lambda_{t,i}(x\mid x_0,x_1)
\log\frac{\lambda_{t,i}(x\mid x_0,x_1)}{\lambda_{\theta,i}(x,t)}
+o(h),
\]
and similarly,
\[
q_h(x-e_i\mid x,x_0,x_1)\log\frac{q_h(x-e_i\mid x,x_0,x_1)}{q_h^{\theta}(x-e_i\mid x,t)}
=
h\,\mu_{t,i}(x\mid x_0,x_1)
\log\frac{\mu_{t,i}(x\mid x_0,x_1)}{\mu_{\theta,i}(x,t)}
+o(h).
\]

Summing all contributions gives
\begin{align*}
% D_{\mathrm{KL}}\!\bigl(q_h(\cdot\mid x,x_0,x_1)\,\Vert\,q_h^{\theta}(\cdot\mid x,t)\bigr)
\mathrm{KL}\!\left(
q_h(\cdot\mid x,x_0,x_1)
\;\middle\|\;
q_h^{\theta}(\cdot\mid x,t)
\right)
&=
h\sum_{i=1}^d
\Biggl[
\lambda_{t,i}(x\mid x_0,x_1)
\log\frac{\lambda_{t,i}(x\mid x_0,x_1)}{\lambda_{\theta,i}(x,t)}
\notag\\
&\qquad\qquad
-\lambda_{t,i}(x\mid x_0,x_1)+\lambda_{\theta,i}(x,t)
\Biggr]
\notag\\
&\quad+
h\sum_{i=1}^d
\Biggl[
\mu_{t,i}(x\mid x_0,x_1)
\log\frac{\mu_{t,i}(x\mid x_0,x_1)}{\mu_{\theta,i}(x,t)}
\notag\\
&\qquad\qquad
-\mu_{t,i}(x\mid x_0,x_1)+\mu_{\theta,i}(x,t)
\Biggr]
+o(h).
\end{align*}
Dividing by $h$ and letting $h\to 0$ yields
\begin{align*}
% \lim_{h\to 0}\frac{1}{h}
% D_{\mathrm{KL}}\!\bigl(q_h(\cdot\mid x,x_0,x_1)\,\Vert\,q_h^{\theta}(\cdot\mid x,t)\bigr)
\lim_{h\to 0}\frac{1}{h}
\mathrm{KL}\!\left(
q_h(\cdot\mid x,x_0,x_1)
\;\middle\|\;
q_h^{\theta}(\cdot\mid x,t)
\right)
&=
\sum_{i=1}^d D_{\mathrm{GKL}}\!\left(\lambda_{t,i}(x\mid x_0,x_1),\lambda_{\theta,i}(x,t)\right) \notag\\
&\quad+
\sum_{i=1}^d D_{\mathrm{GKL}}\!\left(\mu_{t,i}(x\mid x_0,x_1),\mu_{\theta,i}(x,t)\right),
\end{align*}
where $D_{\mathrm{GKL}}$ denotes the generalized KL divergence
\[
D_{\mathrm{GKL}}(u,v)=u\log\frac{u}{v}-u+v.
\]
Up to terms independent of $\theta$, this is exactly
\[
\sum_{i=1}^d\Bigl[\lambda_{\theta,i}(x,t)-\lambda_{t,i}(x\mid x_0,x_1)\log \lambda_{\theta,i}(x,t)\Bigr]
+
\sum_{i=1}^d\Bigl[\mu_{\theta,i}(x,t)-\mu_{t,i}(x\mid x_0,x_1)\log \mu_{\theta,i}(x,t)\Bigr].
\]
% Taking expectation over $(x_0,x_1)$, $t$, and $x$ gives the local KL representation of the training objective up to an additive constant independent of $\theta$.
Taking expectations over $(x_0,x_1)$, $t\sim\mathrm{Unif}[0,\tau]$, and $x$ gives the rate-matching objective in \eqref{eq:rate_matching_objective}, up to an additive constant independent of $\theta$.

\subsection{Path-space KL interpretation of the training objective}
\label{app:path_kl_interp}

% This subsection extends Appendix~\ref{app:poisson_loss} from the local infinitesimal KL between the target and model one-step transition to a path-space KL interpretation of the full training objective.
This subsection integrates the local KL calculation in Appendix~\ref{app:poisson_loss} over time to obtain the path-space KL objective in \eqref{eq:path_kl_objective}.

Let $X_{(t,\tau]}:=\{X_s:\ t<s\le\tau\}$ and define
\[
K_\theta(t)
:=
\mathbb E_{\substack{(x_0,x_1)\sim\pi,\\
x\sim p_t(\cdot\mid x_0,x_1)}}
\left[
\mathrm{KL}\!\left(
p(X_{(t,\tau]}\mid X_t=x,x_0,x_1)
\;\middle\|\;
p_\theta(X_{(t,\tau]}\mid X_t=x)
\right)
\right].
\]

For $0<h\le\tau-t$, the Markov property gives
\begin{equation*}
\begin{aligned}
p(X_{(t,\tau]}\mid X_t=x,x_0,x_1)
&=
p(X_{(t,t+h]}\mid X_t=x,x_0,x_1)\,
p(X_{(t+h,\tau]}\mid X_{t+h},x_0,x_1),\\
p_\theta(X_{(t,\tau]}\mid X_t=x)
&=
p_\theta(X_{(t,t+h]}\mid X_t=x)\,
p_\theta(X_{(t+h,\tau]}\mid X_{t+h}).
\end{aligned}
\end{equation*}
Hence, by the chain rule for KL,
\begin{equation*}
\begin{aligned}
&\mathrm{KL}\!\left(
p(X_{(t,\tau]}\mid X_t=x,x_0,x_1)
\;\middle\|\;
p_\theta(X_{(t,\tau]}\mid X_t=x)
\right)\\
&=
\mathrm{KL}\!\left(
p(X_{(t,t+h]}\mid X_t=x,x_0,x_1)
\;\middle\|\;
p_\theta(X_{(t,t+h]}\mid X_t=x)
\right)\\
&\quad+
\mathbb E_{x'\sim q_h(\cdot\mid x,x_0,x_1)}
\left[
\mathrm{KL}\!\left(
p(X_{(t+h,\tau]}\mid X_{t+h}=x',x_0,x_1)
\;\middle\|\;
p_\theta(X_{(t+h,\tau]}\mid X_{t+h}=x')
\right)
\right].
\end{aligned}
\end{equation*}
After averaging over $(x_0,x_1)\sim\pi$ and
$x\sim p_t(\cdot\mid x_0,x_1)$, the second term is
$K_\theta(t+h)$. Therefore,

\begin{align*}
K_\theta(t)-K_\theta(t+h)
=
\mathbb E_{\substack{(x_0,x_1)\sim\pi,\\
x\sim p_t(\cdot\mid x_0,x_1)}}
\Bigg[
\mathrm{KL}\!\left(
p(X_{(t,t+h]}\mid X_t=x,x_0,x_1)
\;\middle\|\;
p_\theta(X_{(t,t+h]}\mid X_t=x)
\right)
\Bigg].
\end{align*}
Under standard CTMC regularity conditions, the additional path
information within $(t,t+h]$ contributes only $o(h)$ beyond the
endpoint transition. Hence,
\begin{align*}
&\mathrm{KL}\!\left(
p(X_{(t,t+h]}\mid X_t=x,x_0,x_1)
\;\middle\|\;
p_\theta(X_{(t,t+h]}\mid X_t=x)
\right)\\
&\qquad=
\mathrm{KL}\!\left(
q_h(\cdot\mid x,x_0,x_1)
\;\middle\|\;
q_h^\theta(\cdot\mid x,t)
\right)
+
o(h).
\end{align*}

Consequently,
\begin{align*}
K_\theta(t)-K_\theta(t+h)
=
\mathbb E_{\substack{(x_0,x_1)\sim\pi,\\
x\sim p_t(\cdot\mid x_0,x_1)}}
\left[
\mathrm{KL}\!\left(
q_h(\cdot\mid x,x_0,x_1)
\;\middle\|\;
q_h^\theta(\cdot\mid x,t)
\right)
\right]
+
o(h).
\end{align*}

% \textcolor{red}{old}

% For $0<h\le\tau-t$,
% \begin{equation*}
% \begin{aligned}
% p(X_{(t,\tau]}\mid X_t=x,x_0,x_1)
% &=
% q_h(X_{t+h}\mid x,x_0,x_1,t)\,
% p(X_{(t+h,\tau]}\mid X_{t+h},x_0,x_1),\\
% p_\theta(X_{(t,\tau]}\mid X_t=x)
% &=
% q_h^\theta(X_{t+h}\mid x,t)\,
% p_\theta(X_{(t+h,\tau]}\mid X_{t+h}).
% \end{aligned}
% \end{equation*}

% Hence, by the chain rule for KL,
% \begin{equation*}
% \begin{aligned}
% &\mathrm{KL}\!\left(
% p(X_{(t,\tau]}\mid X_t=x,x_0,x_1)
% \;\middle\|\;
% p_\theta(X_{(t,\tau]}\mid X_t=x)
% \right)\\
% &=
% \mathrm{KL}\!\left(
% q_h(\cdot\mid x,x_0,x_1,t)
% \;\middle\|\;
% q_h^\theta(\cdot\mid x,t)
% \right)\\
% &\quad+
% \mathbb E_{x'\sim q_h(\cdot\mid x,x_0,x_1,t)}
% \left[
% \mathrm{KL}\!\left(
% p(X_{(t+h,\tau]}\mid X_{t+h}=x',x_0,x_1)
% \;\middle\|\;
% p_\theta(X_{(t+h,\tau]}\mid X_{t+h}=x')
% \right)
% \right].
% \end{aligned}
% \end{equation*}

% Therefore,
% \[
% K_\theta(t)-K_\theta(t+h)
% =
% \mathbb E_{\substack{(x_0,x_1)\sim\pi,\\
% x\sim p_t(\cdot\mid x_0,x_1)}}
% \left[
% \mathrm{KL}\!\left(
% q_h(\cdot\mid x,x_0,x_1,t)
% \;\middle\|\;
% q_h^\theta(\cdot\mid x,t)
% \right)
% \right].
% \]

% \textcolor{red}{old}

Dividing by $h$ and letting $h\to0$,
\[
-\frac{d}{dt}K_\theta(t)
=
\mathbb E_{\substack{(x_0,x_1)\sim\pi,\\
x\sim p_t(\cdot\mid x_0,x_1)}}
\left[
\lim_{h\to0}\frac{1}{h}
\mathrm{KL}\!\left(
q_h(\cdot\mid x,x_0,x_1,t)
\;\middle\|\;
q_h^\theta(\cdot\mid x,t)
\right)
\right].
\]
By Appendix~\ref{app:poisson_loss}, the expected local rate-matching loss at time $t$ equals
$-\frac{d}{dt}K_\theta(t)$ up to a term independent of $\theta$. Since $K_\theta(\tau)=0$, integrating over $t\in[0,\tau]$ and using the uniform-time objective in \eqref{eq:rate_matching_objective} gives
\[
K_\theta(0)
=
\tau\mathcal L_{\mathrm{train}}(\theta)+c,
\]
where $c$ is independent of $\theta$. By definition,
$K_\theta(0)=\mathcal L_{\mathrm{path}}(\theta)$, which gives the relation stated in Section~\ref{sec:method_training}.

% By Appendix~\ref{app:poisson_loss}, the local rate-matching loss equals
% $-\frac{d}{dt}K_\theta(t)$ up to a term independent of $\theta$. Since $t$ is sampled uniformly on $[0,\tau]$ in \eqref{eq:rate_matching_objective}, integrating over time gives
% \[
% \mathcal L_{\mathrm{path}}(\theta)
% =
% \tau\mathcal L_{\mathrm{train}}(\theta)+c,
% \]
% where $c$ is independent of $\theta$.

% Here $\mathcal L_{\mathrm{path}}(\theta)=K_\theta(0)$ is exactly the path-space KL defined in \eqref{eq:path_kl_objective}.
% \[
% \mathcal{L}_{\mathrm{train}}(\theta)
% =
% \mathbb{E}_{(x_0,x_1)\sim\pi}
% \left[
% \mathrm{KL}\!\left(
% p(X_{(0,\tau]}\mid X_0=x_0,X_1=x_1)
% \;\middle\|\;
% p_\theta(X_{(0,\tau]}\mid X_0=x_0)
% \right)
% \right]
% +
% c.
% \]

\subsection{Error analysis}
\label{app:error_analysis}

In this section, we quantify three sources of error in count-FM, namely endpoint truncation, rate estimation, and sampling discretization. This decomposition parallels error analyses for diffusion samplers that separate endpoint, estimation, and discretization errors \citep{chen2023improved,lu2026mathematical}.
Let $\tau:=1-\varepsilon_t$. Let $p_t^\theta$ denote the marginal distribution at time $t$ of the continuous-time birth-death process with learned rates $(\lambda_\theta,\mu_\theta)$, initialized from $p_0$. 
For a sampling grid $0=t_0<\cdots<t_K=\tau$ with $K$ time intervals and
$\Delta_k:=t_{k+1}-t_k$, let $\widehat p_\tau^\theta$ denote the terminal distribution produced by the discretized sampler in Section~\ref{sec:method_sampling}. 
For distributions $p$ and $q$ on $\mathbb N_0^d$, define the total variation (TV) distance as
$D_{\mathrm{TV}}(p,q):=\frac12\sum_{x\in\mathbb N_0^d}|p(x)-q(x)|$. By the triangle inequality,
\begin{align*}
D_{\mathrm{TV}}(p_1,\widehat p_\tau^\theta)
\le\;&
\underbrace{D_{\mathrm{TV}}(p_1,p_\tau)}_{\text{endpoint truncation error}}
+
\underbrace{D_{\mathrm{TV}}(p_\tau,p_\tau^\theta)}_{\text{rate estimation error}} +
\underbrace{D_{\mathrm{TV}}(p_\tau^\theta,\widehat p_\tau^\theta)}_{\text{discretization error}}.
\end{align*}
We bound these three terms separately below.

\subsubsection{Endpoint truncation error}

For an endpoint pair $(x_0,x_1)\sim\pi$, define $n_i:=|x_1^{(i)}-x_0^{(i)}|$ and $s_i:=\operatorname{sgn}(x_1^{(i)}-x_0^{(i)})$. By the conditional bridge \eqref{eq:conditional_binomial_bridge}, $X_\tau^{(i)}=x_0^{(i)}+s_iB_\tau^{(i)}$, with $B_\tau^{(i)}\sim\mathrm{Binomial}(n_i,\tau)$.
Since $x_1^{(i)}=x_0^{(i)}+s_in_i$, the number of remaining unit jumps satisfies
$$
R_i:=|x_1^{(i)}-X_\tau^{(i)}|
=n_i-B_\tau^{(i)}
\sim\mathrm{Binomial}(n_i,\varepsilon_t).
$$
% Using $(X_\tau,X_1)$ from the same bridge as a coupling, and the coupling characterization
% $D_{\mathrm{TV}}(p,q)=\inf_{\gamma\in\Pi(p,q)}\mathbb P_\gamma(X\neq Y)$, we have
% $D_{\mathrm{TV}}(p_\tau,p_1)\le\mathbb P(X_\tau\neq X_1)$.
Using $(X_\tau,X_1)$ from the same bridge as a coupling, note that for any coupling $(X,Y)$ of $p$ and $q$, $|p(A)-q(A)|\le\mathbb P(X\neq Y)$ for every event $A$. Taking the supremum over $A$ gives the coupling inequality $D_{\mathrm{TV}}(p,q)\le\mathbb P(X\neq Y)$, and hence $D_{\mathrm{TV}}(p_\tau,p_1)\le\mathbb P(X_\tau\neq X_1)$.
Conditionally on $(x_0,x_1)$, reaching $x_1$ requires completing all $n_i$ unit jumps in every coordinate. Since the bridge coordinates are conditionally independent,
\begin{align*}
\mathbb P(X_\tau=X_1\mid x_0,x_1)
&=
\prod_{i=1}^d
\mathbb P(B_\tau^{(i)}=n_i)
=
\prod_{i=1}^d(1-\varepsilon_t)^{n_i}
=
(1-\varepsilon_t)^{\|x_1-x_0\|_1}.
\end{align*}

Therefore, by the law of total expectation,
\begin{align*}
D_{\mathrm{TV}}(p_\tau,p_1)
&\le
\mathbb E_\pi
\left[
1-(1-\varepsilon_t)^{\|X_1-X_0\|_1}
\right]
\le
\varepsilon_t\,\mathbb E_\pi\|X_1-X_0\|_1,
\end{align*}
where the second inequality follows from Bernoulli's inequality,
$1-(1-\varepsilon)^n\le n\varepsilon$.
Thus the endpoint truncation error is $O(\varepsilon_t)$ in total variation.

Although the TV bound above depends on endpoint displacement through the chosen bridge coupling, total variation itself does not account for the geometric distance between count states. 
We therefore also consider a complementary 2-Wasserstein distance $W_2$ bound, which quantifies the magnitude of the remaining displacement in count space \citep{villani2009optimal}. Recall that
$$
W_2^2(p,q)
=
\inf_{\gamma\in\Pi(p,q)}
\mathbb E_{(X,Y)\sim\gamma}\|X-Y\|_2^2.
$$
Assume $\mathbb E_\pi\|X_1-X_0\|_2^2<\infty$.
% Using the same bridge coupling, $W_2^2(p_\tau,p_1)\le\mathbb E\|X_\tau-X_1\|_2^2$.
Since $(X_\tau,X_1)$ is one admissible coupling of $p_\tau$ and $p_1$, the definition of $W_2$ gives $W_2^2(p_\tau,p_1)\le\mathbb E\|X_\tau-X_1\|_2^2$.

Since $|X_\tau^{(i)}-X_1^{(i)}|=R_i$ and, for each coordinate $i$,
$$
\mathbb E[R_i^2\mid x_0,x_1]
=
n_i\varepsilon_t(1-\varepsilon_t)
+
n_i^2\varepsilon_t^2,
$$
we have
\begin{align*}
\mathbb E\!\left[
\|X_\tau-X_1\|_2^2
\mid x_0,x_1
\right]
&=
\sum_{i=1}^d
\mathbb E[R_i^2\mid x_0,x_1] \\
&=
\varepsilon_t(1-\varepsilon_t)\|x_1-x_0\|_1
+
\varepsilon_t^2\|x_1-x_0\|_2^2.
\end{align*}
Averaging over $\pi$ gives
\begin{align*}
W_2^2(p_\tau,p_1)
\le\;&
\varepsilon_t(1-\varepsilon_t)
\mathbb E_\pi\|X_1-X_0\|_1 +
\varepsilon_t^2
\mathbb E_\pi\|X_1-X_0\|_2^2.
\end{align*}
Hence $W_2(p_\tau,p_1)=O(\sqrt{\varepsilon_t})$ for fixed endpoint moments. The quantity $\|X_1-X_0\|_1$ is exactly the number of unit jumps required by the conditional bridge, showing how the source distribution and endpoint coupling affect truncation difficulty.

\subsubsection{Rate estimation error}

Let $(\lambda_t,\mu_t)$ denote the exact marginal transition rates introduced in Section~\ref{sec:method_training}. Define
$$
\mathcal E_{\mathrm{rate}}
=
\int_0^\tau
\mathbb E_{X_t\sim p_t}
\sum_{i=1}^d
\Big[
D_{\mathrm{GKL}}\!\left(
\lambda_{t,i}(X_t),
\lambda_{\theta,i}(X_t,t)
\right) +
D_{\mathrm{GKL}}\!\left(
\mu_{t,i}(X_t),
\mu_{\theta,i}(X_t,t)
\right)
\Big]dt,
$$
where $D_{\mathrm{GKL}}$ is defined in Appendix~\ref{app:poisson_loss}. 
% Under standard integrability and absolute-continuity conditions, applying the CTMC path-space KL argument \citep{arampatzis2015pathwise} of Appendix~\ref{app:path_kl_interp} to the exact and learned marginal processes and then using data processing gives
% $\mathrm{KL}(p_\tau\Vert p_\tau^\theta)\le\mathcal E_{\mathrm{rate}}$.
% Pinsker's inequality \citep{bakry2014analysis} therefore gives
Let $\mathbb P$ and $\mathbb P_\theta$ denote the path distributions of the exact and learned marginal CTMCs on $[0,\tau]$. Under standard integrability and absolute-continuity conditions, the path-space KL argument \citep{arampatzis2015pathwise} of Appendix~\ref{app:path_kl_interp} gives $\mathrm{KL}(\mathbb P\Vert\mathbb P_\theta)=\mathcal E_{\mathrm{rate}}$. 
Since $p_\tau$ and $p_\tau^\theta$ are the corresponding distributions at time $\tau$, the data-processing inequality gives $\mathrm{KL}(p_\tau\Vert p_\tau^\theta)\le \mathrm{KL}(\mathbb P\Vert\mathbb P_\theta)=\mathcal E_{\mathrm{rate}}$.
Pinsker's inequality \citep{bakry2014analysis} therefore gives
$$
D_{\mathrm{TV}}(p_\tau,p_\tau^\theta)
\le
\sqrt{\frac12
\mathrm{KL}(p_\tau\Vert p_\tau^\theta)} \le
\sqrt{\frac12\,\mathcal E_{\mathrm{rate}}}.
$$
We next relate $\mathcal E_{\mathrm{rate}}$ to the training objective. Let $\mathcal L_{\mathrm{train}}^*$ denote the value of \eqref{eq:rate_matching_objective} when the model rates are replaced by the exact marginal rates $(\lambda_t,\mu_t)$. By definition, the marginal rate averages the conditional bridge rates over endpoint pairs given the current state,
$
\lambda_{t,i}(X_t)
=
% \mathbb E
\mathbb E_{\substack{(X_0,X_1)\sim\pi,
X_t\sim p_t(\cdot\mid X_0,X_1)}}
\!\left[
\lambda_{t,i}(X_t\mid X_0,X_1)
\mid X_t
\right],
$
and analogously for the death rates. Since $\ell(u,v)=v-u\log v$ is affine in $u$,
$$
\mathbb E\!\left[
\ell\!\left(
\lambda_{t,i}(X_t\mid X_0,X_1),v
\right)
\mid X_t
\right]
=
v-
\mathbb E\!\left[
\lambda_{t,i}(X_t\mid X_0,X_1)
\mid X_t
\right]\log v
=
\ell\!\left(
\lambda_{t,i}(X_t),v
\right).
$$
Thus, after averaging over endpoint pairs, the population conditional rate-matching objective is equivalent to matching the exact marginal rates.

Recall that the rate-matching loss in \eqref{eq:rate_matching_objective} is
$\ell(u,v)=v-u\log v$.
For a marginal target rate $u$ and model rate $v$, the excess loss over the oracle choice $v=u$ is
$$
\ell(u,v)-\ell(u,u)
=
u\log\frac{u}{v}-u+v
=
D_{\mathrm{GKL}}(u,v).
$$
Therefore, summing this excess loss over all birth and death coordinates, averaging over $X_t\sim p_t$, and integrating over $t\in[0,\tau]$ gives
$\tau\left[
\mathcal L_{\mathrm{train}}(\theta)-\mathcal L_{\mathrm{train}}^*
\right]
=
\mathcal E_{\mathrm{rate}}$.
Combining this identity with the TV bound above gives
$$
D_{\mathrm{TV}}(p_\tau,p_\tau^\theta)
\le
\sqrt{
\frac{\tau}{2}
\left[
\mathcal L_{\mathrm{train}}(\theta)
-
\mathcal L_{\mathrm{train}}^*
\right]
}.
$$
Thus minimizing the excess rate-matching loss controls not only local transition-rate error but also the terminal distribution error of the learned continuous-time process.

\subsubsection{Discretization error}

For coordinate $i$, recall
$r_i(x,t)=\lambda_{\theta,i}(x,t)+\mu_{\theta,i}(x,t)$
from Section~\ref{sec:method_sampling}, and define the total jump rate
$a_t^\theta(x,t)=\sum_{i=1}^d r_i(x,t)$ (Appendix~\ref{app:poisson_loss}).
Also define the rate vector
$\nu_\theta(x,t):=(\lambda_\theta(x,t),\mu_\theta(x,t))$.
Assume $a_t^\theta(x,t)\le A$ and
$$
\|\nu_\theta(y,s)-\nu_\theta(x,t)\|_1
\le
L_x\|y-x\|_1+L_t|s-t|
$$
for all $x,y\in\mathbb N_0^d$ and $s,t\in[0,\tau]$, where $L_x$ and $L_t$ are the Lipschitz constants for state and time, respectively.

% For coordinate $i$, recall
% $r_i(x,t)=\lambda_{\theta,i}(x,t)+\mu_{\theta,i}(x,t)$
% from Section~\ref{sec:method_sampling}, and define the total jump rate
% $a_t^\theta(x,t)=\sum_{i=1}^d r_i(x,t)$ (Appendix~\ref{app:poisson_loss}).
% Assume $a_t^\theta(x,t)\le A$ and, for all $x,y\in\mathbb N_0^d$ and $s,t\in[0,\tau]$,

% \begin{align*}
% \sum_{i=1}^d
% \Big[
% &|\lambda_{\theta,i}(y,s)-\lambda_{\theta,i}(x,t)|
% +
% |\mu_{\theta,i}(y,s)-\mu_{\theta,i}(x,t)|
% \Big]
% \le
% L_x\|y-x\|_1+L_t|s-t|,
% \end{align*}

% where $L_x$ and $L_t$ are the Lipschitz constants for state and time, respectively.

Let $P_k^\theta(x,\cdot)$ denote the exact learned CTMC transition over $[t_k,t_{k+1}]$ starting from $x$, and let $\widehat P_k^\theta(x,\cdot)$ denote the corresponding transition of the implemented local-jump sampler. To separate the two sources of one-step discretization error, define a frozen-rate transition over $[t_k,t_{k+1}]$. Starting from $x$ at $t_k$, the rates evaluated at $(x,t_k)$ are held constant throughout the interval. If $z$ denotes the current state within the interval $[t_k,t_{k+1}]$,
$$
\bar\lambda_i(z)=\lambda_{\theta,i}(x,t_k),
\qquad
\bar\mu_i(z)=\mu_{\theta,i}(x,t_k)\mathbf 1\{z^{(i)}>0\}.
$$

Let $\bar P_k^\theta(x,\cdot)$ denote the resulting transition distribution. Then
$$
D_{\mathrm{TV}}
\left(
P_k^\theta(x,\cdot),
\widehat P_k^\theta(x,\cdot)
\right)
\le\;
\underbrace{
D_{\mathrm{TV}}
\left(
P_k^\theta(x,\cdot),
\bar P_k^\theta(x,\cdot)
\right)
}_{\text{rate-freezing error}} +
\underbrace{
D_{\mathrm{TV}}
\left(
\bar P_k^\theta(x,\cdot),
\widehat P_k^\theta(x,\cdot)
\right)
}_{\text{multiple-jump error}}.
$$
We bound these two terms separately below.

\paragraph{Multiple-jump error.}
The implemented sampler allows at most one jump per coordinate during $[t_k,t_{k+1}]$, while the frozen-rate process may make multiple jumps. Let $N_i$ denote the number of actual jumps in coordinate $i$, and let
$\widetilde N_i\sim\mathrm{Poisson}(r_i(x,t_k)\Delta_k)$
denote the number of candidate events under the constant frozen rate. Since death events are suppressed at zero, $N_i$ is stochastically dominated by $\widetilde N_i$.

% The frozen process and the sampler agree whenever every coordinate makes at most one jump. 
To bound $D_{\mathrm{TV}}(\bar P_k^\theta(x,\cdot),\widehat P_k^\theta(x,\cdot))$, the second term above, note that the frozen process and the sampler agree whenever every coordinate makes at most one jump.
Therefore,
\begin{align*}
D_{\mathrm{TV}}
\left(
\bar P_k^\theta(x,\cdot),
\widehat P_k^\theta(x,\cdot)
\right)
&\le
\mathbb P\!\left(\bigcup_{i=1}^d\{N_i\ge2\}\right)
\le
\sum_{i=1}^d\mathbb P(\widetilde N_i\ge2) \\
&\le
\frac12\sum_{i=1}^d
\mathbb E\!\left[\widetilde N_i(\widetilde N_i-1)\right]
=
\frac{\Delta_k^2}{2}
\sum_{i=1}^d r_i(x,t_k)^2,
\end{align*}
where $\mathbb P(N\ge2)\le\mathbb E[N(N-1)]/2=m^2/2$ for $N\sim\mathrm{Poisson}(m)$.

Thus the local multiple-jump error is $O(\Delta_k^2)$ and scales with the squared jump-rate sum,
$\sum_{i=1}^d r_i(x,t_k)^2$. The bound therefore suggests using smaller time steps $\Delta_k$ in regions where the squared jump-rate sum is large.

\paragraph{Rate-freezing error.}
To bound $D_{\mathrm{TV}}(P_k^\theta(x,\cdot),\bar P_k^\theta(x,\cdot))$, the first term above, 
let $Y_u$ denote the exact learned process after elapsed time $u\in[0,\Delta_k]$ from $t_k$, with $Y_0=x$. 
Comparing the exact and frozen-rate processes over the same interval, their TV difference is bounded by the accumulated difference in their jump rates. Applying the Lipschitz bound,
\begin{align*}
D_{\mathrm{TV}}
\left(
P_k^\theta(x,\cdot),
\bar P_k^\theta(x,\cdot)
\right)
&\le
\int_0^{\Delta_k}
\mathbb E
\left[
\|\nu_\theta(Y_u,t_k+u)-\nu_\theta(x,t_k)\|_1
\right]du \\
&\le
\int_0^{\Delta_k}
\left[
L_tu+L_x\,\mathbb E\|Y_u-x\|_1
\right]du \\
&\le
\frac{\Delta_k^2}{2}(L_t+A L_x).
\end{align*}
The last inequality uses $a_t^\theta\le A$, where $a_t^\theta(x,t)=\sum_{i=1}^d r_i(x,t)$. Since the CTMC only makes local transitions $x\to x\pm e_i$, its displacement from $x$ is bounded by the number of jumps, giving $\mathbb E\|Y_u-x\|_1\le Au$. Thus the local rate-freezing error is $O(\Delta_k^2)$ and grows with $L_t+A L_x$.

% \paragraph{Rate-freezing error.}
% Let $Y_u$ denote the exact learned process after elapsed time $u\in[0,\Delta_k]$ from $t_k$, with $Y_0=x$. A coupling of the exact and frozen-rate processes gives

% \begin{align*}
% D_{\mathrm{TV}}
% \left(
% P_k^\theta(x,\cdot),
% \bar P_k^\theta(x,\cdot)
% \right)
% &\le
% \int_0^{\Delta_k}
% \mathbb E
% \sum_{i=1}^d
% \Big[
% |\lambda_{\theta,i}(Y_u,t_k+u)-\bar\lambda_i(Y_u)|
% +
% |\mu_{\theta,i}(Y_u,t_k+u)-\bar\mu_i(Y_u)|
% \Big]du \\
% &\le
% \int_0^{\Delta_k}
% \left[
% L_tu+L_x\,\mathbb E\|Y_u-x\|_1
% \right]du \\
% &\le
% \int_0^{\Delta_k}
% \left[
% L_tu+A L_xu
% \right]du
% =
% \frac{\Delta_k^2}{2}(L_t+A L_x).
% \end{align*}

% The second inequality follows from the Lipschitz assumption. Since each birth or death changes the $L_1$ displacement by at most one and the total jump rate is bounded by $A$, we have $\mathbb E\|Y_u-x\|_1\le Au$. Thus the local rate-freezing error is $O(\Delta_k^2)$ and increases with $L_t+A L_x$, suggesting smaller $\Delta_k$ when the learned rates vary rapidly in time or state.

Combining the two contributions,
\begin{align*}
D_{\mathrm{TV}}
\left(
P_k^\theta(x,\cdot),
\widehat P_k^\theta(x,\cdot)
\right)
\le
\frac{\Delta_k^2}{2}
\left[
\sum_{i=1}^d r_i(x,t_k)^2
+
L_t+A L_x
\right].
\end{align*}

Thus the one-step discretization error is $O(\Delta_k^2)$. Using TV contraction under Markov kernels and summing over the $K$ intervals,
\begin{align*}
D_{\mathrm{TV}}
(p_\tau^\theta,\widehat p_\tau^\theta)
\le
\frac12
\sum_{k=0}^{K-1}
\Delta_k^2
\left[
\mathbb E_{X_{t_k}\sim p_{t_k}^\theta}
\sum_{i=1}^d r_i(X_{t_k},t_k)^2
+
L_t+A L_x
\right].
\end{align*}

For the uniform grid used in Section~\ref{sec:method_sampling},
$\Delta_k=\Delta=\tau/K$. Using the global rate bound,
$\sum_i r_i^2\le(\sum_i r_i)^2\le A^2$,
$$
D_{\mathrm{TV}}(p_\tau^\theta,\widehat p_\tau^\theta)
\le
\frac{\tau^2}{2K}
\left(
A^2+L_t+A L_x
\right)
=
O(K^{-1}).
$$

Thus the sampler is globally first order. The local error increases with the squared jump-rate sum $\sum_{i=1}^d r_i(x,t_k)^2$ and the rate-variation term $L_t+A L_x$, suggesting the use of smaller time steps $\Delta_k$ when either term is large.

\subsubsection{Overall error bound}

Combining the endpoint truncation, rate estimation, and uniform-grid discretization bounds gives
\begin{align*}
D_{\mathrm{TV}}(p_1,\widehat p_\tau^\theta)
\le\;&
\underbrace{
\varepsilon_t\,\mathbb E_\pi\|X_1-X_0\|_1
}_{\text{endpoint truncation}}
\\
&+
\underbrace{
\sqrt{
\frac{\tau}{2}
\left[
\mathcal L_{\mathrm{train}}(\theta)
-
\mathcal L_{\mathrm{train}}^*
\right]
}
}_{\text{rate estimation}}
\\
&+
\underbrace{
\frac{\tau^2}{2K}
\left(
A^2+L_t+A L_x
\right)
}_{\text{discretization}}.
\end{align*}
Thus the overall terminal error separates into an $O(\varepsilon_t)$ endpoint term, a term controlled by the excess rate-matching loss, and an $O(K^{-1})$ discretization term. If the endpoint moments and the rate bounds $A$, $L_x$, and $L_t$ remain bounded, then the bound vanishes as $\varepsilon_t\to0$, $\mathcal L_{\mathrm{train}}(\theta)-\mathcal L_{\mathrm{train}}^*\to0$, and $K\to\infty$.

\section{Simulation}
\label{sim_appendix}

Unless otherwise stated, all experiments were run on a single NVIDIA RTX 4090 GPU with 24GB memory.

\begin{figure}[h!]
  \centering
  \includegraphics[width=0.8\linewidth]{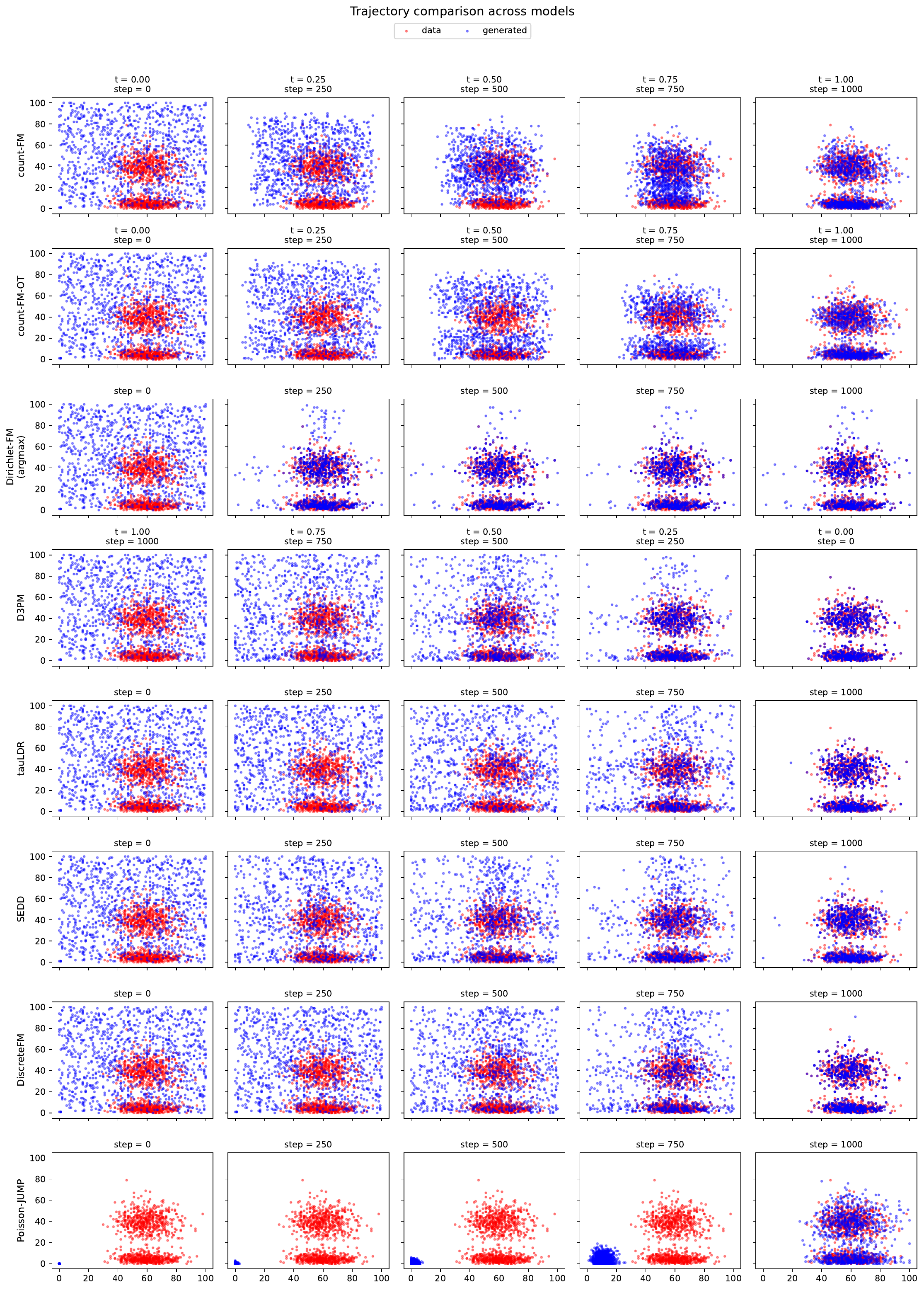}
  \caption{\textbf{Representative intermediate samples from the native sampling trajectories of different models in the simulation.} Red points show target samples and blue points show generated samples. Columns correspond to increasing native sampling time or sampling step within each method's own generation procedure.}
  \label{fig:sim_snapshot}
\end{figure}

Figure~\ref{fig:sim_snapshot} visualizes representative intermediate samples along each model's native sampling trajectory in the simulation. Since different methods use different native time parameterizations and different sampling mechanisms, this figure is qualitative and is intended only to show how generated samples evolve under each fitted model's own dynamics.

\begin{figure}[h!]
  \centering
  \includegraphics[width=0.8\linewidth]{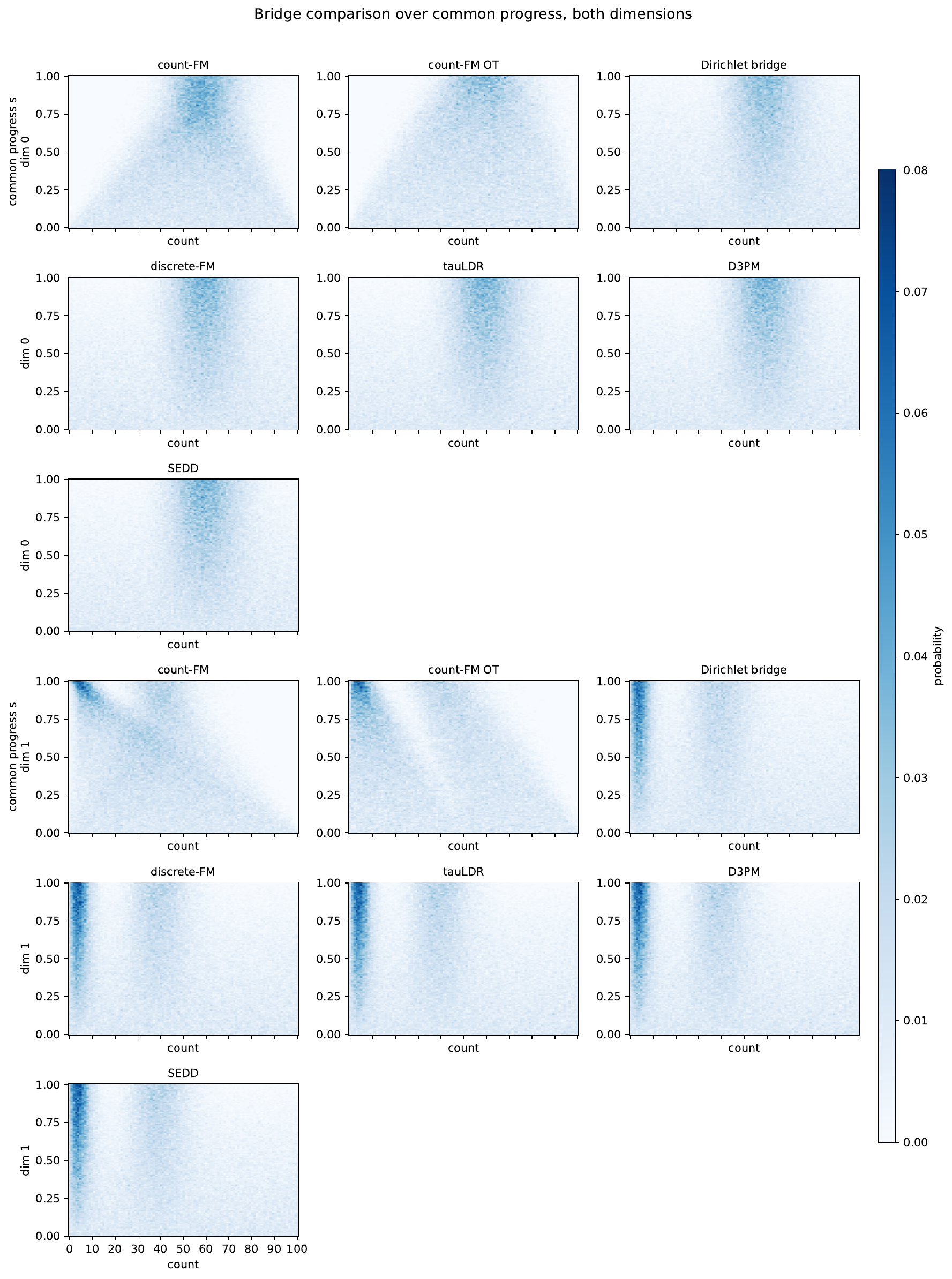}
  \caption{\textbf{Marginal bridge distributions for both coordinates.} They are shown under a common progress variable $s$. Intermediate marginals are estimated by Monte Carlo sampling from each model's bridge or forward noising distribution after mapping to the common progress scale. Count-FM evolves gradually across $s$, while categorical-state baselines become target-like early. count-FM-OT uses the same bridge as count-FM but with OT-coupled endpoints, yielding a visibly straighter transition path.}
  \label{fig:sim_model_bridge}
\end{figure}

To compare intermediate bridge behavior across models, we construct, for each method and each coordinate, an empirical family of one-dimensional intermediate marginals indexed by a common progress variable $s\in[0,1]$. For a fixed value of $s$, we repeatedly sample endpoint pairs $(x_0^{(m)},x_1^{(m)})$, generate an intermediate state $X_s^{(m)}$ using the corresponding model-specific bridge or forward noising distribution, and estimate the marginal pmf by
\[
\widehat p_s^{(i)}(z)=\frac{1}{M}\sum_{m=1}^M \mathbf{1}\!\left\{X_s^{(i,m)}=z\right\},
\qquad z\in\{0,1,\dots,C_{\max}\}.
\]
The heatmaps in Figure~\ref{fig:sim_model_bridge} plot $\widehat p_s^{(i)}(z)$ as a function of count value $z$ and common progress $s$.

For count-FM, the conditional bridge is
\[
X_t^{(i)} = x_0^{(i)} + \operatorname{sgn}\!\left(x_1^{(i)}-x_0^{(i)}\right) B_t^{(i)},
\qquad
B_t^{(i)} \sim \mathrm{Binomial}\!\left(\left|x_1^{(i)}-x_0^{(i)}\right|, t\right),
\]
and we set $s=t$. For count-FM-OT, we use the same conditional bridge, but the endpoints $(x_0,x_1)$ are first paired using the minibatch OT coupling from Section~\ref{sec:method_ot}. 
We do not include Poisson-JUMP in Figure~\ref{fig:sim_model_bridge}, because its Poisson-source jump construction does not provide a directly comparable one-parameter bridge or forward noising family under the same common-progress normalization.

For the remaining baselines, since each method uses a different forward process and parameterization, we convert each native time variable to a common progress variable $s\in[0,1]$ by monotone rescaling of a model-specific progress statistic $\rho(t)$,
\[
s=\frac{\rho(t)-\rho(0)}{\rho(1)-\rho(0)}.
\]
Here $\rho(t)$ measures source-to-target progress along the native path of each method.
% For discrete-FM, we use its native interpolation variable. For D3PM, tauLDR, and SEDD, we use the retained-source-signal quantity defined by the corresponding forward process. For the Dirichlet bridge, we use the probability assigned to the target category along the bridge. 
This places all comparison methods on the same source-to-target scale. Under this normalization, the count-FM bridges evolve gradually across the full range of $s$, whereas the categorical-state diffusion- and flow-based baselines typically become target-like much earlier, reflecting a sharper transition in count space.

\section{OT coupling and sampling efficiency}
\label{app:ot_efficiency_appendix}

\begin{figure}[h!]
  \centering
  \includegraphics[width=\linewidth]{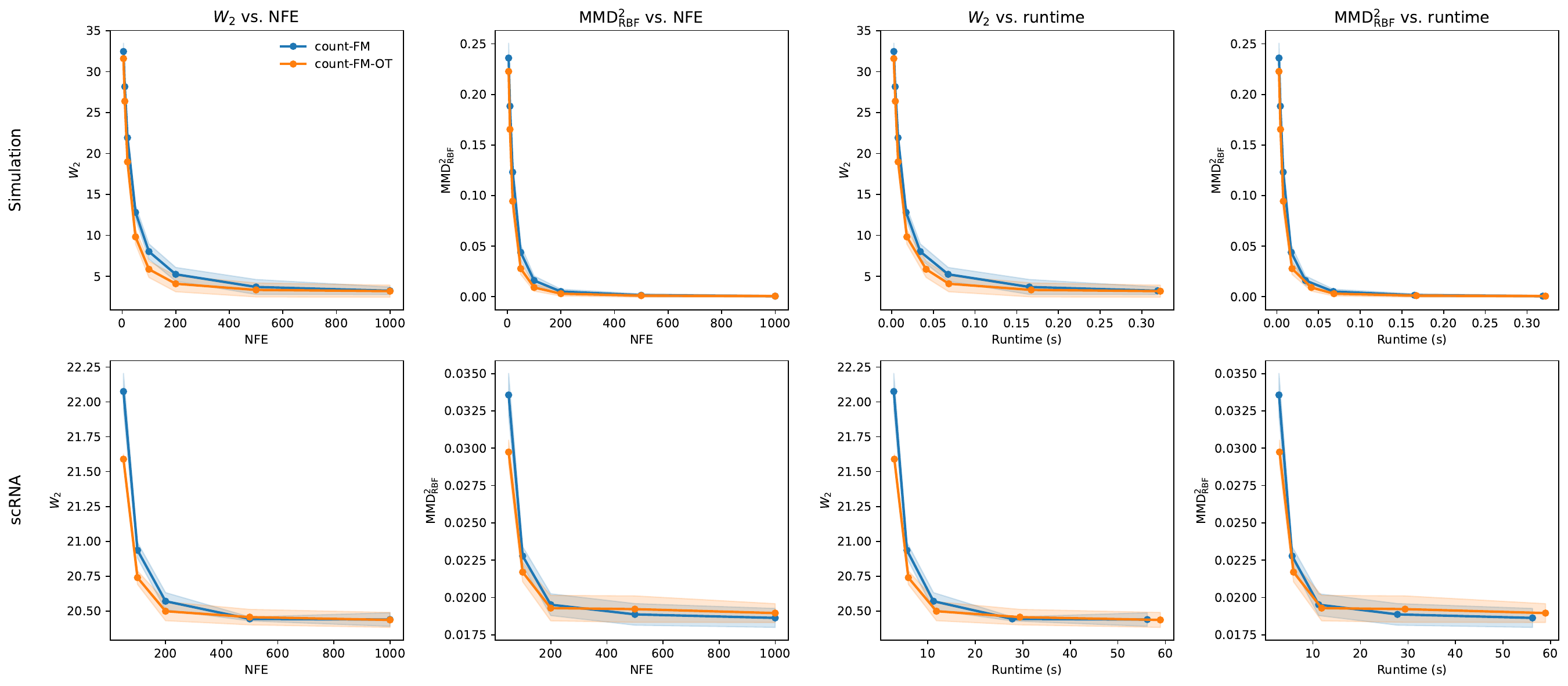}
  \caption{\textbf{Sampling efficiency comparison between count-FM and count-FM-OT in the simulation and scRNA experiments.} We plot $W_2$ and $\mathrm{MMD}^2_{\mathrm{RBF}}$ against the number of function evaluations (NFE) and wall-clock runtime. In the simulation, count-FM-OT reaches a given quality level with smaller computational budget. On the scRNA task, the same trend is present but weaker, and the difference becomes small at larger budgets.}
  \label{fig:ot_efficiency}
\end{figure}

Figure~\ref{fig:ot_efficiency} compares the sampling efficiency of count-FM and count-FM-OT in both the simulation and scRNA experiments. We plot $W_2$ and $\mathrm{MMD}^2_{\mathrm{RBF}}$ against the number of function evaluations (NFE) and wall-clock runtime. In the simulation, count-FM-OT shows a clear efficiency advantage across most computational budgets. On the scRNA task, the advantage is smaller but still visible at lower budgets, while the two methods become very similar as the budget increases.

\section{Applications}
\label{app_appendix}

\subsection{single-cell RNA-seq generation and transport}
\label{app:scRNA_appendix}

For unconditional generation, we perform paired comparisons across five matched generation seeds. Differences are baseline minus count-FM, so positive values favor count-FM. We report the paired standardized effect size $d_z=\overline{\Delta}/s_\Delta$ and Holm-corrected $p$-values \citep{holm1979simple}.
\begin{table}[h!]
  \caption{\textbf{Paired comparison with count-FM on the Dentate Gyrus dataset.}}
  \label{tab:scrna_paired}
  \centering
  \small
  \begin{tabular}{lrrrrrr}
    \toprule
    Baseline &
    $\Delta W_2$ & $d_z$ & Holm $p$ &
    $\Delta \mathrm{MMD}^2_{\mathrm{RBF}}$ & $d_z$ & Holm $p$ \\
    \midrule
    scDiffusion & 0.0886 & 1.27 & 0.0466
                & 0.00401 & 6.23 & $3.39\times10^{-4}$ \\
    scLDM       & 0.2823 & 7.20 & $2.61\times10^{-4}$
                & 0.00175 & 2.47 & 0.00525 \\
    scVI        & 0.2303 & 3.91 & 0.00188
                & 0.00439 & 6.74 & $3.39\times10^{-4}$ \\
    DCM         & 3.1095 & 50.26 & $1.88\times10^{-7}$
                & 0.01111 & 21.50 & $5.60\times10^{-6}$ \\
    CFGen       & 1.7700 & 21.98 & $4.10\times10^{-6}$
                & 0.01766 & 16.89 & $1.17\times10^{-5}$ \\
    \bottomrule
  \end{tabular}
\end{table}

Although some gains are modest, all differences remain significant after Holm correction.

PCA-space $W_2$ and $\mathrm{MMD}^2_{\mathrm{RBF}}$ measure global distributional agreement rather than biological fidelity directly. We therefore select marker genes using the training data. For each annotated cell type, we retain the 10 genes with the largest log-normalized mean-expression difference from all other cell types, yielding 99 unique markers.
\begin{table}[h!]
  \caption{\textbf{Marker-gene recovery on the Dentate Gyrus dataset.}}
  \label{tab:scrna_marker}
  \centering
  \begin{tabular}{lcc}
    \toprule
    Model
    & Marker correlation $\uparrow$
    & Marker MAE $\downarrow$ \\
    \midrule
    count-FM
    & \textbf{0.9959}
    & \textbf{0.0652} \\

    scLDM
    & 0.9941
    & 0.0697 \\

    scDiffusion
    & 0.9927
    & 0.0710 \\

    scVI
    & 0.9934
    & 0.0778 \\

    DCM
    & 0.9710
    & 0.1296 \\

    CFGen
    & 0.9926
    & 0.1534 \\
    \bottomrule
  \end{tabular}
\end{table}

\begin{figure}[h!]
  \centering
  \includegraphics[width=0.8\linewidth]{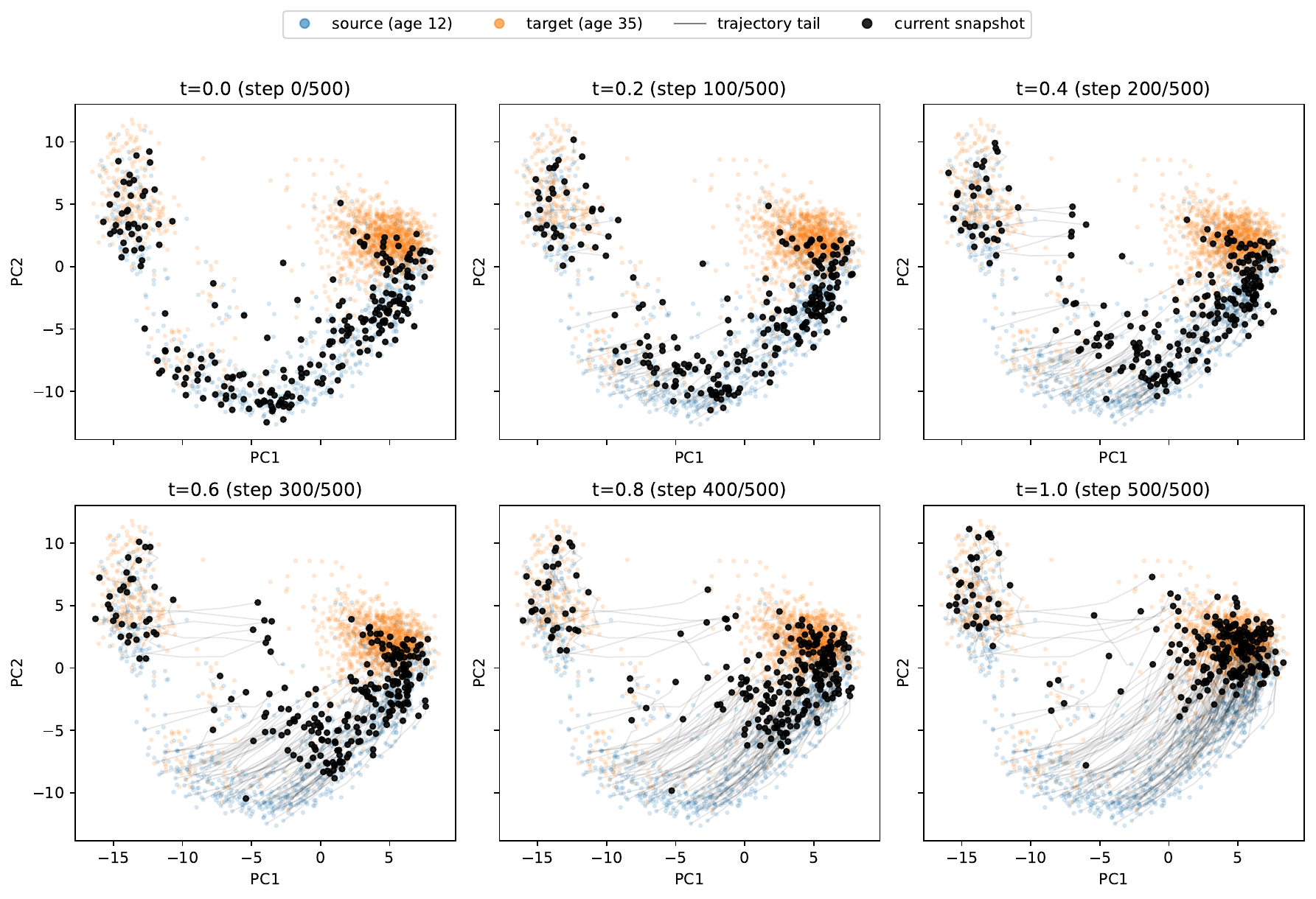}
  \caption{\textbf{Transition snapshots for training.} As time increases, generated cells move progressively from the P12 source manifold toward the P35 target manifold in PCA space. Light colored points show the source and target references, gray curves show trajectory tails, and black points show the current generated snapshot.}
  \label{fig:scRNA_snapshot_train}
\end{figure}

\begin{figure}[h!]
  \centering
  \includegraphics[width=0.8\linewidth]{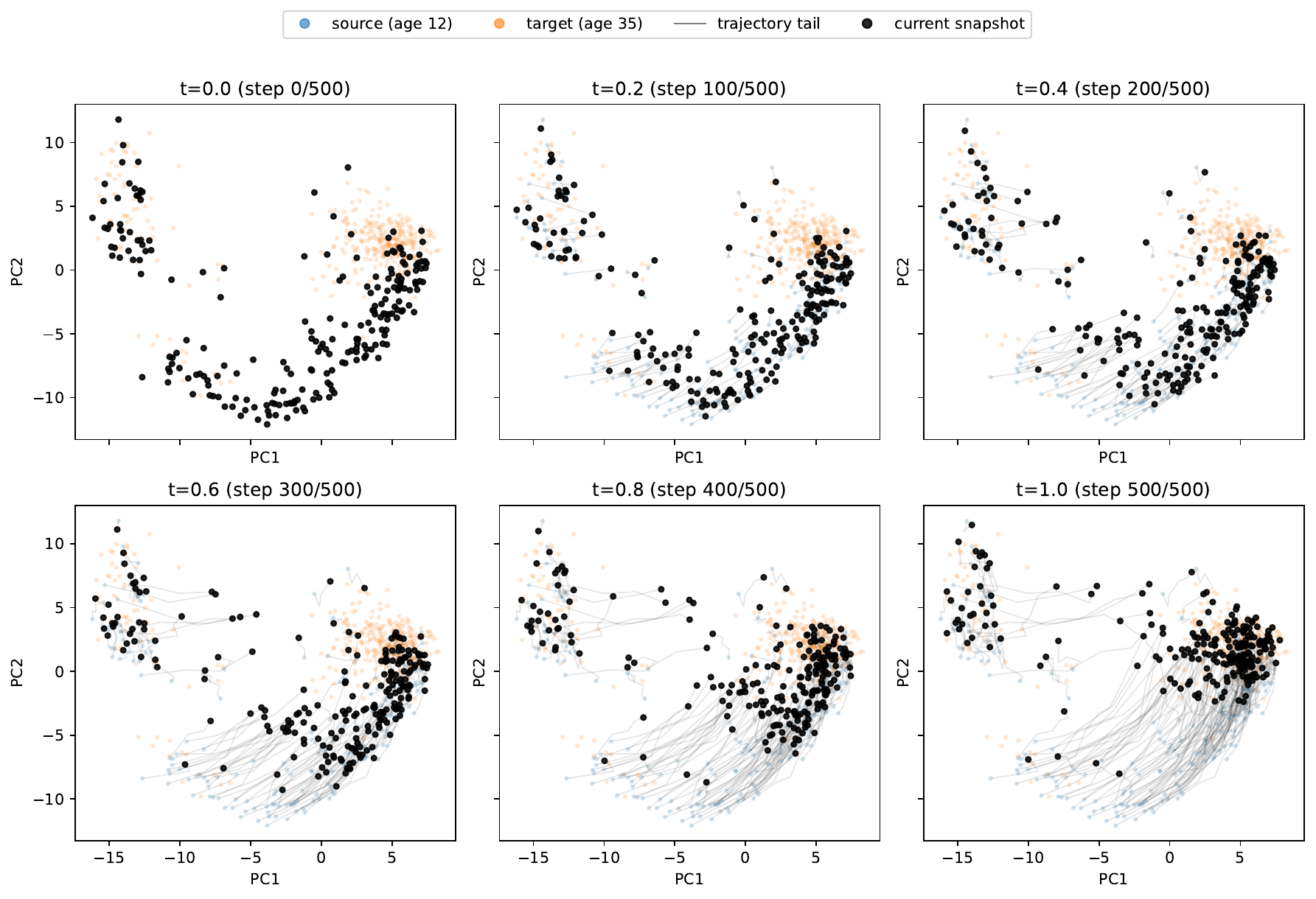}
  \caption{\textbf{Transition snapshots for testing.} The same progressive transport pattern is observed on held-out cells, with trajectories moving smoothly from the P12 manifold toward the P35 manifold.}
  \label{fig:scRNA_snapshot_test}
\end{figure}

All methods achieve high marker-expression correlations, while count-FM obtains the lowest marker MAE, indicating the most accurate recovery of cell-type-relevant expression levels. Appendix Figures~\ref{fig:scRNA_snapshot_train} and~\ref{fig:scRNA_snapshot_test} complement the main-text trajectory and fate summaries by visualizing the temporal evolution of the transport process on both the training and test sets.

\subsection{Additional results for hippocampal hc-3}
\label{app:hc3_appendix}

\begin{figure}[h!]
  \centering
  \includegraphics[width=\linewidth]{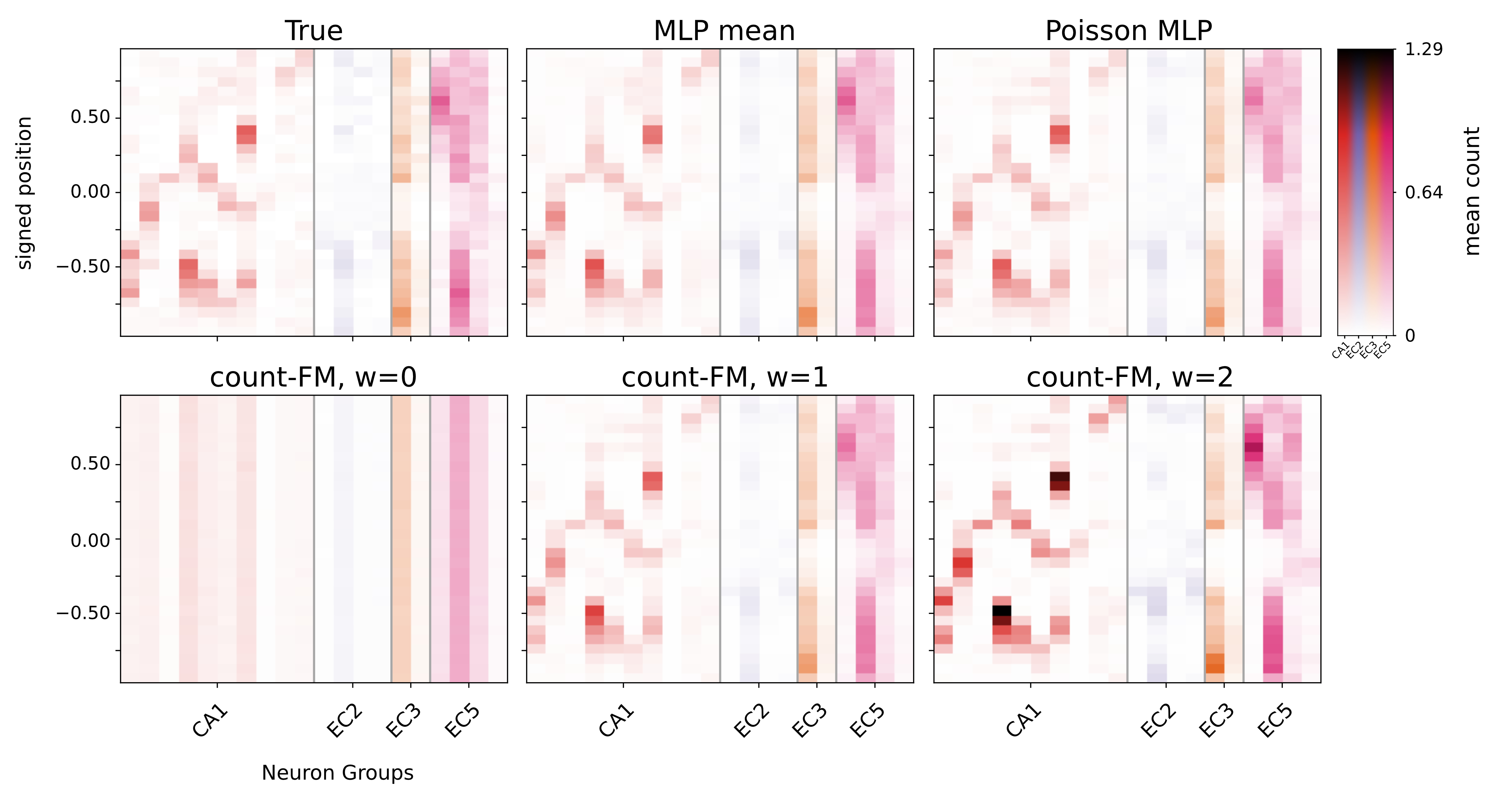}
  \caption{\textbf{Mean-response comparison for the hc-3 conditional-generation task.} Each panel shows bin-wise mean counts across signed position, with neurons grouped by region. For generative models, summaries are estimated from 100 generated samples per held-out covariate. The MLP mean regressor, Poisson MLP, and count-FM with $w=1$ recover the broad spatial response pattern. Count-FM with $w=0$ lacks spatial specificity, while $w=2$ sharpens location-specific responses but reduces calibration.}
  \label{fig:hc3_mean_heatmap}
\end{figure}

This appendix provides expanded qualitative diagnostics for the hc-3 conditional-generation experiment. For generative models, all plotted summaries are estimated from 100 generated samples per held-out covariate. Appendix Figure~\ref{fig:hc3_mean_heatmap} shows bin-wise mean responses across signed position for all models, with neurons grouped by brain region. The MLP mean regressor, Poisson MLP, and count-FM with $w=1$ all recover the broad spatial response pattern. Count-FM with $w=0$ lacks conditional information and only captures marginal firing-rate differences across neurons, while $w=2$ sharpens location-specific responses but is less well calibrated to the true mean pattern.

\begin{figure}[h!]
  \centering
  \includegraphics[width=0.95\linewidth]{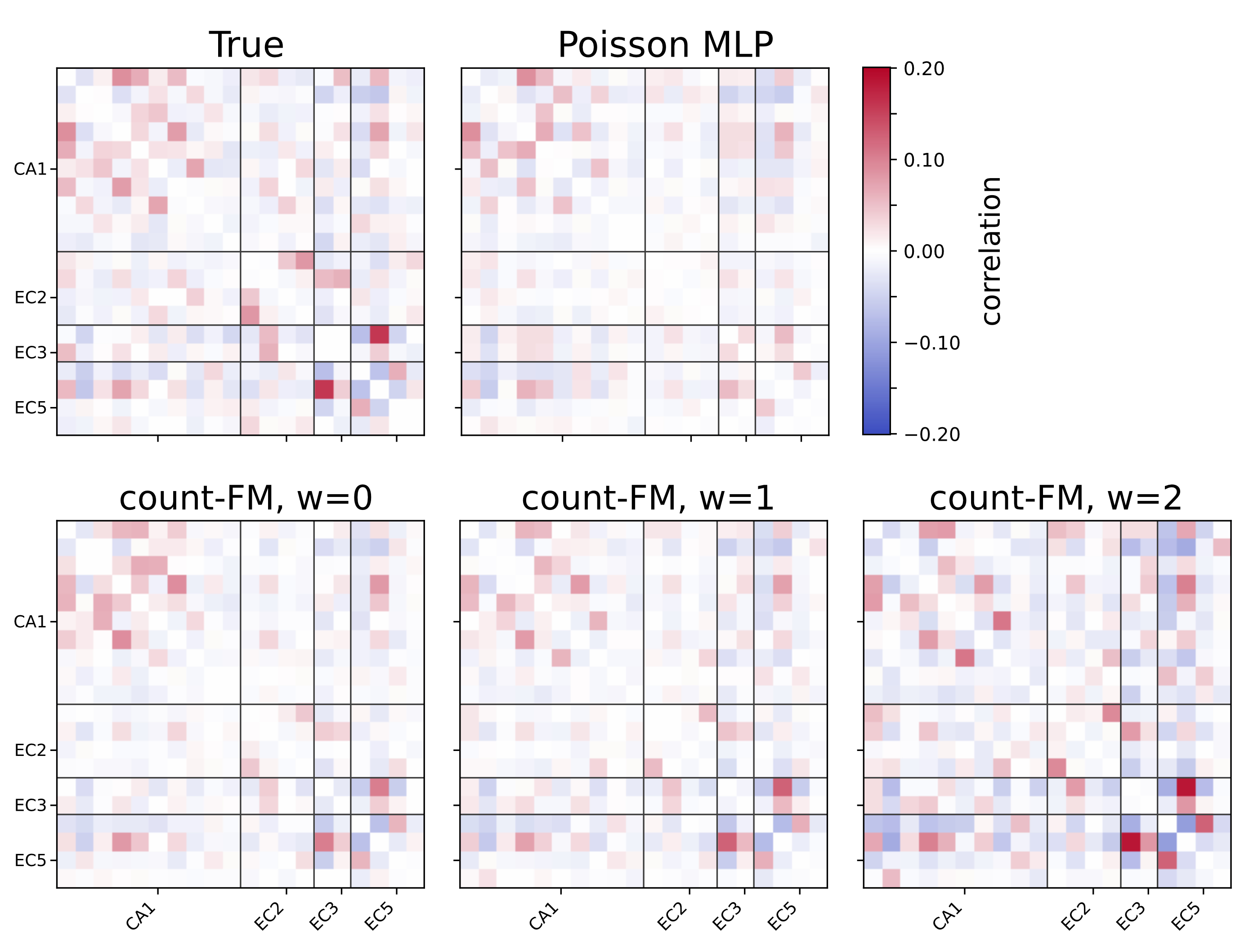}
  \caption{\textbf{Population correlation comparison for the hc-3 conditional-generation task.} Panels show correlation matrices for the active-neuron set. These are estimated from 100 generated samples per held-out covariate. Count-FM with $w=1$ most closely matches the true dependence structure, while the Poisson MLP underestimates population correlations. Count-FM with $w=0$ still captures some marginal dependence structure, and stronger guidance ($w=2$) amplifies the correlation pattern but is less well calibrated.}
  \label{fig:hc3_dependence}
\end{figure}

Appendix Figure~\ref{fig:hc3_dependence} shows the corresponding population correlation structure using the same neuron set. Since the MLP mean regressor is deterministic and is not a generative count model, it is omitted from this dependence comparison. Count-FM with $w=1$ most closely matches the true dependence structure, while the Poisson MLP underestimates population correlations. The unconditional model ($w=0$) still captures some marginal dependence structure, whereas $w=2$ amplifies the correlation pattern but is less well calibrated. These are consistent with Table~\ref{tab:hc3_conditional}, where $w=1$ gives the best overall generative fidelity, whereas $w=2$ mainly sharpens the response pattern at the cost of calibration.

\subsection{Conditional generation of piriform cortex spike trains}
\label{app:odor_spike}

We also evaluate count-FM on conditional generation of multichannel piriform cortex (PCx) spike trains using the public odor-response dataset of \citet{BoldingFranks2017,eLife53125}. The dataset contains processed, spike-sorted extracellular recordings together with simultaneously recorded respiration traces from head-fixed mice. We focus on the PCx recording from session 141208-2 (bank 2), which contains 480 trials. Each trial is aligned to inhalation onset, spikes are binned into 10 ms intervals over the window $[-0.5,1.5]$ s, and each sample is represented as a time-by-neuron count matrix with 200 time bins and 84 neurons \citep{Miura2012OdorRepresentations}. The conditioning variables include time, a respiration-waveform covariate, odor presence, odor identity, and binarized inhalation state. We use a trial-level 80/20 train-test split stratified by odor identity, giving 384 training trials and 96 held-out test trials.

\begin{figure}[h!]
  \centering
  \includegraphics[width=\linewidth]{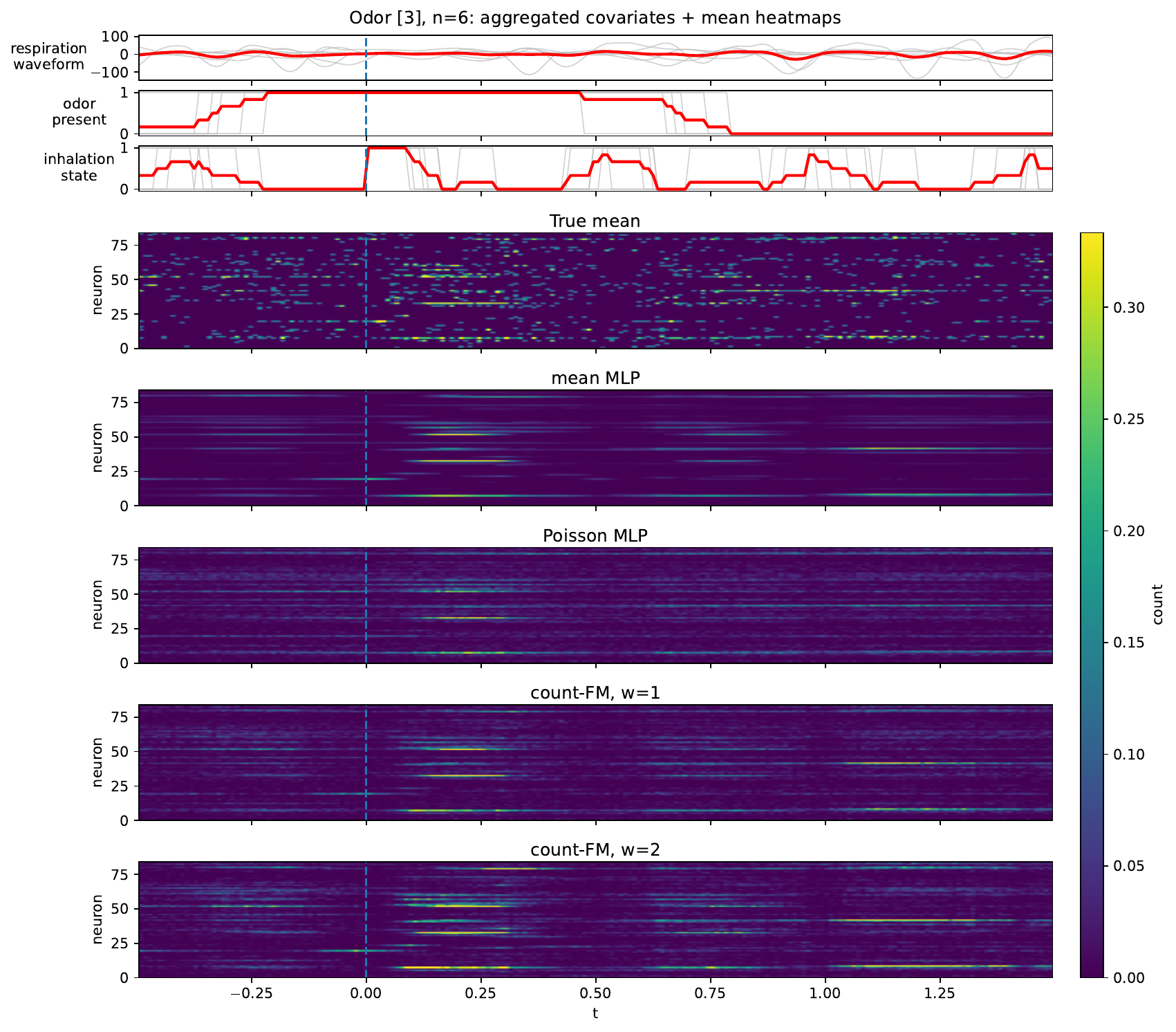}
  \caption{\textbf{Conditional generation on held-out odor-3 trials.} Upper panel shows conditioning covariates over time, with gray traces for individual held-out trials and red traces for trial averages. Lower panel shows neuron-by-time mean responses for the held-out data, the MLP mean regressor, Poisson MLP, and count-FM with $w=1$ and $w=2$. The MLP mean regressor is overly smooth and Poisson MLP remains diffuse, while count-FM better recovers localized odor- and respiration-dependent responses. Increasing guidance to $w=2$ sharpens the response pattern but over-amplifies spike counts, reducing amplitude calibration.}
  \label{fig:odor_spike_mean_spike}
\end{figure}

\begin{figure}[h!]
  \centering
  \includegraphics[width=\linewidth]{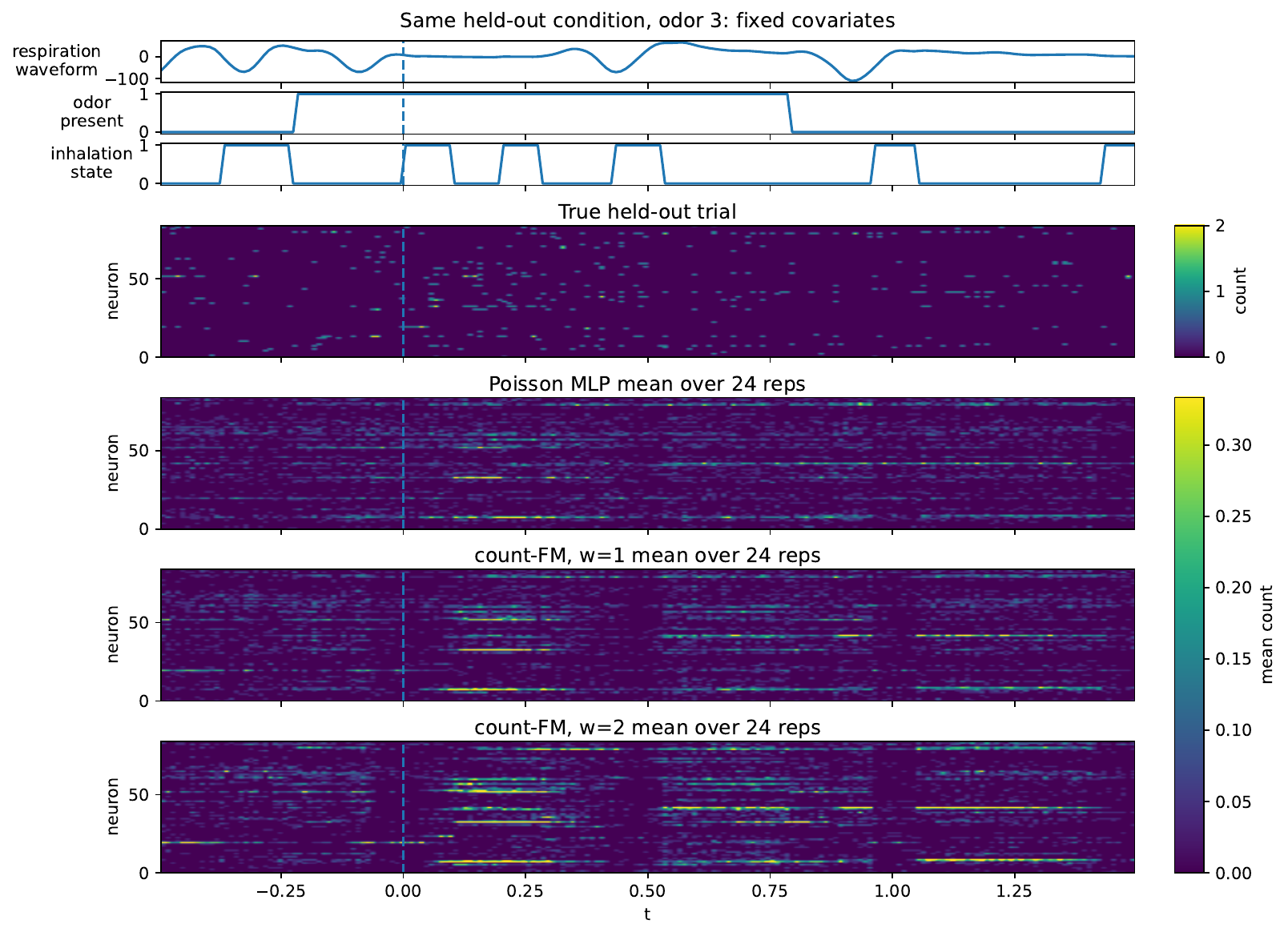}
  \caption{\textbf{Generated responses under the same held-out covariate trajectory.} We compare the true held-out trial with the mean response from the Poisson MLP regressor over 24 repetitions, and with the count-FM mean responses over 24 repetitions at $w=1$ and $w=2$. Compared with the Poisson MLP regressor, count-FM produces a more structured condition-aligned response. Increasing guidance from $w=1$ to $w=2$ sharpens the shape further, but also inflates the response amplitude.}
  \label{fig:odor_spike_fixed_cond}
\end{figure}

Figure~\ref{fig:odor_spike_mean_spike} shows mean responses over held-out odor-3 trials. The mean MLP captures the coarse response shape but is smoother than the held-out mean, while the Poisson MLP is a stronger count baseline but remains comparatively diffuse. Count-FM recovers the odor- and respiration-dependent structure better. Increasing guidance from $w=1$ to $w=2$ sharpens the aligned response pattern, but also increases its amplitude. After aligning the spiking activity with covariates (upper panel in Figure~\ref{fig:odor_spike_mean_spike}) by time, we can see that the neural activity is strongly correlated with the inhalation state (lower firing rate during inhalation), especially when the odor is present. Figure~\ref{fig:odor_spike_fixed_cond} shows the same guidance tradeoff under a fixed held-out covariate. Here, the mean MLP is not shown because it is not generative.

To quantify what we observed, we calculate different metrics and Table~\ref{tab:odor_spike_summary} summarizes held-out metrics over 5 replicated runs, each with a new odor-stratified 80/20 train-test split and retraining under different random seeds. Among the count-FM settings, $w=1$ gives the best overall tradeoff, with the lowest mean RMSE and 95th-percentile error, while remaining close to the Poisson MLP baseline in matched cosine and mean-count ratio. By contrast, $w=2$ gives the highest matched cosine among count-FM settings, but substantially worse mean RMSE and a larger mean-count ratio, indicating stronger condition alignment at the cost of amplitude distortion.

\begin{table}[h!]
 \caption{\textbf{Held-out conditional generation quality on the PCx task.} Results are mean $\pm$ standard deviation over 5 random seeds, each with a new odor-stratified train-test split and model retraining. count-FM with $w=1$ gives the best overall tradeoff among count-FM settings, while $w=2$ improves matched cosine but over-amplifies spike counts.}
  \label{tab:odor_spike_summary}
  \centering
  \small
  \begin{tabular}{lcccc}
    \toprule
    Model & Mean RMSE $\downarrow$ & Matched cosine $\uparrow$ & Mean-count ratio & 95th-percentile error $\downarrow$ \\
    \midrule
    MLP mean          & 1.264 $\pm$ 0.248 & \textbf{0.800 $\pm$ 0.041} & 0.704 $\pm$ 0.100 & 0.487 $\pm$ 0.017 \\
    Poisson MLP       & 1.090 $\pm$ 0.071 & 0.762 $\pm$ 0.018 & 1.073 $\pm$ 0.058 & 0.311 $\pm$ 0.069 \\
    count-FM, $w=0$   & 2.090 $\pm$ 0.255 & 0.407 $\pm$ 0.074 & 1.300 $\pm$ 0.429 & 0.322 $\pm$ 0.093 \\
    count-FM, $w=1$   & \textbf{1.082 $\pm$ 0.056} & 0.760 $\pm$ 0.019 & \textbf{1.037 $\pm$ 0.078} & \textbf{0.297 $\pm$ 0.071} \\
    count-FM, $w=2$   & 1.744 $\pm$ 0.342 & \underline{0.792 $\pm$ 0.022} & 1.427 $\pm$ 0.188 & 0.336 $\pm$ 0.029 \\
    \bottomrule
  \end{tabular}
\end{table}

% \begin{figure}[h!]
%   \centering
%   \includegraphics[width=\linewidth]{figures/pcx_appendix_neuron_marginal_scatter_with_poisson.pdf}
%   \caption{Neuron-wise marginal diagnostics on held-out PCx data. Each point compares held-out and generated statistics for one neuron. The top row shows neuron-wise mean firing rate, and the bottom row shows zero fraction, for the Poisson MLP regressor and count-FM at $w=0$, $1$, and $2$. Count-FM with $w=1$ remains closer to the diagonal, while $w=2$ overshoots neuron means and reduces zero fractions.}
%   \label{fig:odor_spike_diagnostic}
% \end{figure}

%%%%%%%%%%%%%%%%%%%%%%%%%%%%%%%%%%%%%%%%%%%%%%%%%%%%%%%%%%%%
\clearpage
\newpage

\section*{NeurIPS Paper Checklist}

\begin{enumerate}

\item {\bf Claims}
    \item[] Question: Do the main claims made in the abstract and introduction accurately reflect the paper's contributions and scope?
    \item[] Answer: \answerYes{} % Replace by \answerYes{}, \answerNo{}, or \answerNA{}.
    \item[] Justification: The abstract and introduction summarize the proposed method, experiments, and scope of the claims.

\item {\bf Limitations}
    \item[] Question: Does the paper discuss the limitations of the work performed by the authors?
    \item[] Answer: \answerYes{} % Replace by \answerYes{}, \answerNo{}, or \answerNA{}.
    \item[] Justification: Limitations and future directions are discussed in Section~\ref{discussion}.

\item {\bf Theory assumptions and proofs}
    \item[] Question: For each theoretical result, does the paper provide the full set of assumptions and a complete (and correct) proof?
    \item[] Answer: \answerYes{} % Replace by \answerYes{}, \answerNo{}, or \answerNA{}.
    \item[] Justification: Derivations of the bridge rates and KL objectives are provided in Appendix~\ref{app:method_derivations}.

    \item {\bf Experimental result reproducibility}
    \item[] Question: Does the paper fully disclose all the information needed to reproduce the main experimental results of the paper to the extent that it affects the main claims and/or conclusions of the paper (regardless of whether the code and data are provided or not)?
    \item[] Answer: \answerYes{} % Replace by \answerYes{}, \answerNo{}, or \answerNA{}.
    \item[] Justification: Datasets, metrics and model settings are described in the main text and appendix.

\item {\bf Open access to data and code}
    \item[] Question: Does the paper provide open access to the data and code, with sufficient instructions to faithfully reproduce the main experimental results, as described in supplemental material?
    \item[] Answer: \answerYes{} % Replace by \answerYes{}, \answerNo{}, or \answerNA{}.
    \item[] Justification: The experiments use public datasets and code is attached as Supplementary Material.

\item {\bf Experimental setting/details}
    \item[] Question: Does the paper specify all the training and test details (e.g., data splits, hyperparameters, how they were chosen, type of optimizer) necessary to understand the results?
    \item[] Answer: \answerYes{} % Replace by \answerYes{}, \answerNo{}, or \answerNA{}.
    \item[] Justification: All reported.

\item {\bf Experiment statistical significance}
    \item[] Question: Does the paper report error bars suitably and correctly defined or other appropriate information about the statistical significance of the experiments?
    \item[] Answer: \answerYes{} % Replace by \answerYes{}, \answerNo{}, or \answerNA{}.
    \item[] Justification: See tables in the main text.

\item {\bf Experiments compute resources}
    \item[] Question: For each experiment, does the paper provide sufficient information on the computer resources (type of compute workers, memory, time of execution) needed to reproduce the experiments?
    \item[] Answer: \answerYes{} % Replace by \answerYes{}, \answerNo{}, or \answerNA{}.
    \item[] Justification: provided in the appendix.
    
\item {\bf Code of ethics}
    \item[] Question: Does the research conducted in the paper conform, in every respect, with the NeurIPS Code of Ethics \url{https://neurips.cc/public/EthicsGuidelines}?
    \item[] Answer: \answerYes{} % Replace by \answerYes{}, \answerNo{}, or \answerNA{}.
    \item[] Justification: public datasets and does not involve new human-subject data collection.

\item {\bf Broader impacts}
    \item[] Question: Does the paper discuss both potential positive societal impacts and negative societal impacts of the work performed?
    \item[] Answer: \answerNo{} % Replace by \answerYes{}, \answerNo{}, or \answerNA{}.
    \item[] Justification: This is a methodological paper.
    
\item {\bf Safeguards}
    \item[] Question: Does the paper describe safeguards that have been put in place for responsible release of data or models that have a high risk for misuse (e.g., pre-trained language models, image generators, or scraped datasets)?
    \item[] Answer: \answerNA{} % Replace by \answerYes{}, \answerNo{}, or \answerNA{}.
    \item[] Justification: no high-risk models or datasets.

\item {\bf Licenses for existing assets}
    \item[] Question: Are the creators or original owners of assets (e.g., code, data, models), used in the paper, properly credited and are the license and terms of use explicitly mentioned and properly respected?
    \item[] Answer: \answerYes{} % Replace by \answerYes{}, \answerNo{}, or \answerNA{}.
    \item[] Justification: Public dataset.

\item {\bf New assets}
    \item[] Question: Are new assets introduced in the paper well documented and is the documentation provided alongside the assets?
    \item[] Answer: \answerNA{} % Replace by \answerYes{}, \answerNo{}, or \answerNA{}.
    \item[] Justification: The paper does not release a new dataset or benchmark asset.

\item {\bf Crowdsourcing and research with human subjects}
    \item[] Question: For crowdsourcing experiments and research with human subjects, does the paper include the full text of instructions given to participants and screenshots, if applicable, as well as details about compensation (if any)? 
    \item[] Answer: \answerNA{} % Replace by \answerYes{}, \answerNo{}, or \answerNA{}.
    \item[] Justification: no crowdsourcing or new human-subject experiments.

\item {\bf Institutional review board (IRB) approvals or equivalent for research with human subjects}
    \item[] Question: Does the paper describe potential risks incurred by study participants, whether such risks were disclosed to the subjects, and whether Institutional Review Board (IRB) approvals (or an equivalent approval/review based on the requirements of your country or institution) were obtained?
    \item[] Answer: \answerNA{} % Replace by \answerYes{}, \answerNo{}, or \answerNA{}.
    \item[] Justification: no human-subject experiments.

\item {\bf Declaration of LLM usage}
    \item[] Question: Does the paper describe the usage of LLMs if it is an important, original, or non-standard component of the core methods in this research? Note that if the LLM is used only for writing, editing, or formatting purposes and does \emph{not} impact the core methodology, scientific rigor, or originality of the research, declaration is not required.
    %this research? 
    \item[] Answer: \answerNA{} % Replace by \answerYes{}, \answerNo{}, or \answerNA{}.
    \item[] Justification: LLMs are not part of the core method or experiments.

\end{enumerate}

\end{document}